\documentclass[10pt,twocolumn,letterpaper]{article}

\usepackage[pagenumbers]{cvpr} 

\usepackage{graphicx}
\usepackage{amsmath}
\usepackage{amssymb}
\usepackage{booktabs}
\usepackage{bm}
\usepackage{subcaption}
\usepackage{multirow}
\usepackage{bbding}

\usepackage[pagebackref,breaklinks,colorlinks]{hyperref}
\usepackage[capitalize]{cleveref}

\crefname{section}{Sec.}{Secs.}
\Crefname{section}{Section}{Sections}
\Crefname{table}{Table}{Tables}
\crefname{table}{Tab.}{Tabs.}

\begin{document}

\title{LAB-Net: LAB Color-Space Oriented Lightweight Network \\ for Shadow Removal}

\author{Hong Yang$^1$\footnotemark[1], Gongrui Nan$^1$\footnotemark[1], Mingbao Lin$^{2}$, Fei Chao$^1$\footnotemark[2], Yunhang Shen$^2$, Ke Li$^2$, Rongrong Ji$^{1}$\\
$^1$MAC Lab, School of Informatics, Xiamen University \quad
$^2$Tencent Youtu Lab  \\
\normalfont{\{yanghong, nangongrui\}@stu.xmu.edu.cn\quad
linmb001@outlook.com \quad
fchao@xmu.edu.cn } \\
\normalfont{
\{shenyunhang01, tristanli.sh\}@gmail.com \quad
rrji@xmu.edu.cn}}

\maketitle

\renewcommand{\thefootnote}{\fnsymbol{footnote}} 
\footnotetext[1]{Indicates equal contribution}
\footnotetext[2]{Corresponding author}

\begin{abstract}
This paper focuses on the limitations of current over-parameterized shadow removal models. We present a novel lightweight deep neural network that processes shadow images in the LAB color space. The proposed network termed ``LAB-Net'', is motivated by the following three observations:
First, the LAB color space can well separate the luminance information and color properties.
Second, sequentially-stacked convolutional layers fail to take full use of features from different receptive fields.
Third, non-shadow regions are important prior knowledge to diminish the drastic color difference between shadow and non-shadow regions.
Consequently, we design our LAB-Net by involving a two-branch structure: L and AB branches. Thus the shadow-related luminance information can well be processed in the L branch, while the color property is well retained in the AB branch.
In addition, each branch is composed of several Basic Blocks,  local spatial attention modules (LSA), and convolutional filters. Each Basic Block consists of multiple parallelized dilated convolutions of divergent dilation rates to receive different receptive fields that are operated with distinct network widths to save model parameters and computational costs. Then, an enhanced channel attention module (ECA) is constructed to aggregate features from different receptive fields for better shadow removal.
Finally, the LSA modules are further developed to fully use the prior information in non-shadow regions to cleanse the shadow regions. 
We perform extensive experiments on the both ISTD and SRD datasets. 
Experimental results show that our LAB-Net well outperforms state-of-the-art methods. Also, our model's parameters and computational costs are reduced by several orders of magnitude. Our code is available at \url{https://github.com/ngrxmu/LAB-Net}.
\end{abstract}

\section{Introduction}
\label{sec:intro}
Shadows, widely existing in various natural scenes, are caused when light sources are fully or partially occluded by objects, which greatly abate the image quality and further degrade the performance of many downstream tasks such as object detection~\cite{cucchiara2003detecting}, object tracking~\cite{sanin2010improved,guo2020spark}, \emph{etc}. Therefore, how to accurately remove image shadows has received long-time attention in the artificial intelligence research community, and yet remains unsolved.

To obtain shadow-free images, current Deep Neural Networks (DNNs)-based methods~\cite{qu2017deshadownet,wang2018stacked,cun2020towards,liu2021shadow} learned mappings between shadow and non-shadow images under deep learning frameworks. 
Albeit the promising performance, we realize in this paper that three important factors including \textit{inappropriate color representation}, \textit{unitary receptive field} and \textit{underutilized non-shadow prior}, are sorely neglected in most existing methods so that the progress of shadow removal remains stagnant.

In terms of inappropriate color representation, most modern shadow removal networks processed images in the RGB color space~\cite{wang2018stacked,cun2020towards,zhu2022efficient} without considering the luminance information, with which the shadow removal task is highly correlated. 
%
Instead, it is natural to choose a representation including the luminance information. The LAB color space (composed of L, A, and B channels) has aroused great interest in the research community since the perceptual luminance is separately stored in the L channel while the A and B channels contain color information.
~\cite{hu2019direction} simply optimized all the L, A, and B channels as a whole entity. However, it leads to limited performance since the luminance and color properties are intermingled.
In addition, ~\cite{liu2021shadow_b} chose to process the L channel first, then the processed L channel is used to guide the learning of both the A and B channels.
However, solely processing the L channel results in a sub-optimal result, which in turn affects the learning of the A and B channels.

For the unitary receptive field issue,  most current networks for shadow removal consist of sequentially-stacked convolutions with fixed-sized receptive fields to process incoming per-layer features~\cite{hu2019mask,chen2021canet,fu2021auto}. However,  the fixed-sized receptive fields cannot fully consider multi-scale context information,  so as to limit the performance gains. Therefore, it is very necessary to import convolution operations of different receptive fields. This inspired many researchers to integrate dilated convolutions of divergent dilation rates to expand the receptive fields without introducing additional parameters. For example, DHAN~\cite{cun2020towards} concatenated a series of dilated convolutions with a linearly growing dilation rate to capture multi-scale context information.
However, the dilated convolutions are arranged in a tandem fashion, so as to cause that different features are not fully explored. In addition, the computational costs are not well decreased since the dilated convolutions with various dilation rates use an identical number of channels to process inputs.

In terms of underutilized non-shadow prior information, most current implementations on shadow removal~\cite{wang2018stacked,cun2020towards,fu2021auto} directly learned to cleanse shadow regions while underestimating the efficacy of non-shadow regions. The non-shadow regions indeed contain valuable prior information since their luminance information and color property are very close to the ground truth of shadow regions. Ignoring this fact leads to color inconsistencies between shadow and non-shadow regions in images generated by existing methods. Therefore, excavating suitable prior information in non-shadow regions has become particularly important. 
For example, \cite{chen2021canet} introduced a pre-trained contextual patch matching module to generate a set of potential matching pairs of shadow and non-shadow patches. Nevertheless, the pre-trained module not only heavily relies on a pre-defined dataset, but also leads to the great complexity of the entire network pipeline.

Therefore in this paper, we believe all the above three issues are interrelated and can be addressed in a unified framework. Similar to~\cite{hu2019direction,liu2021shadow_b}, we deal with shadow-contaminated images in the LAB color space and accordingly develop a novel lightweight shadow removal network, termed LAB-Net as depicted in Fig.\,\ref{fig:LAB-Net}. Different from the existing studies, where L, A, and B channels are processed as a whole entity~\cite{hu2019direction} or alternately~\cite{liu2021shadow_b}, our LAB-Net applies a two-branch structure to process the luminance channel L and color channels A \& B in parallel. Then, these two branches interact with each other for information exchange, allowing their optimization to respectively reach the optimal.

In order to detect context information in multiple scales, we also consider dilated convolutions. However, in contrast to~\cite{cun2020towards}, as illustrated in Fig.\,\ref{fig:Basic block}, the Basic Block of our LAB-Net is composed of multiple parallelized dilated convolutions with divergent dilation rates in each layer. Such a structure provides the feasibility of allocating computational costs to context information of different scales. Therefore, the complexity of our LAB-Net can be well reduced. Besides, a Laplacian filter-based enhanced channel attention module (ECA) is deployed at the latter part of each Basic Block, so that the multi-scale features from different levels can be well aggregated to better cleanse shadow images.

Finally, to make full use of non-shadow prior information, we further establish a local spatial attention module (LSA), whose structure is shown in Fig.\,\ref{fig:LSA}. Our LSA focuses on non-shadow pixels around the spatial boundaries of shadow regions and constructs a correlation matrix between non-shadow and shadow regions to assist image recovery processes. In contrast to the big-budget matching module~\cite{chen2021canet}, the spatial locality in our LSA circumvents accessing all non-shadow pixels, resulting in few computational costs.

We conduct extensive experiments on two widely-used shadow removal datasets including ISTD~\cite{wang2018stacked} and SRD~\cite{qu2017deshadownet}. Experimental results show that our LAB-Net mostly manifests its great merits in better performance and lower complexity over many existing state-of-the-art methods.

\section{Related Work}
\label{sec:related}
In this section, we briefly revisit existing studies that are mostly related to our work, including literature on shadow removal as well as discussions on the RGB and LAB image color representations.

\subsection{Shadow Removal}
There are two mainstreams in the shadow removal field including traditional physics-based methods and modern DNNs-based methods.
Physics-based methods eliminate shadows by utilizing inherent prior knowledge such as illumination, spatial correlation, \emph{etc}.
~\cite{shor2008shadow,yang2012shadow,xiao2013fast,zhang2015shadow,zhang2018improving} proposed to restore shadow images from prior knowledge of illumination. ~\cite{guo2012paired} removed shadows by using spatial characteristics of both shadow and non-shadow regions.
The boosting of DNNs in the past decades has also led to substantial breakthroughs in shadow removal. 
~\cite{qu2017deshadownet} for the first time built a multi-context deep neural network to remove shadows, while~\cite{cun2020towards} proposed a two-layer aggregation network and synthesized realistic shadow images through adversarial learning. 
~\cite{fu2021auto} formulated the shadow removal problem as a task to automatically fuse multi-exposure images. ~\cite{le2019shadow} built a shadow illumination model and a particularly designed network to remove shadows. Furthermore, ~\cite{zhu2022efficient} introduced iterative optimization to improve the interpretability of the model as well as the effect of shadow removal. Inspired by the attention mechanism, several recent studies explored pixel correlations within an image.
\begin{figure*}[!th]
    \centering
    \includegraphics[width=1.0\linewidth]{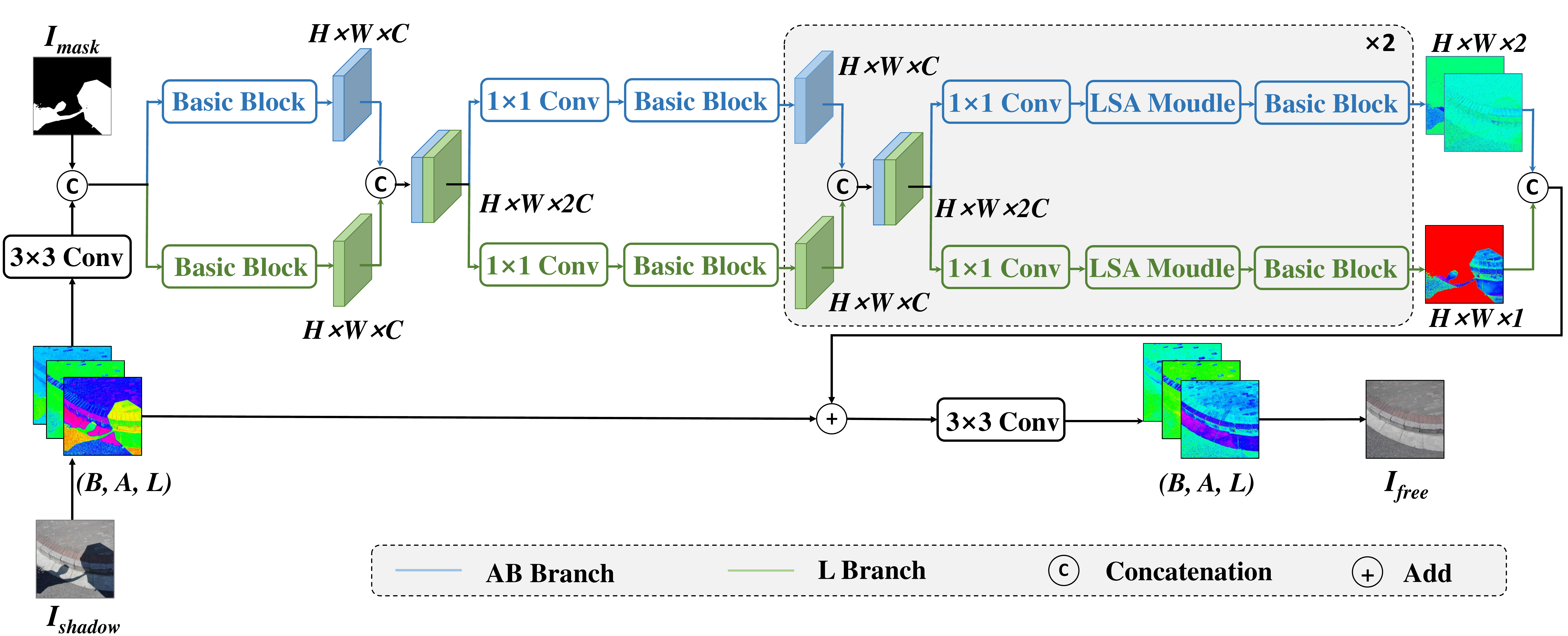}
    \caption{Overall framework of LAB-Net.}
    \label{fig:LAB-Net}
\end{figure*}
~\cite{hu2019direction} used direction-aware spatial context information to improve shadow detection and shadow removal. %
~\cite{chen2021canet} adopted the contextual feature transfer mechanism to transfer non-shadow region features to similar shadow regions to detect contextual matching information. %

\subsection{RGB \emph{v.s.} LAB Representation}
Most image processing tasks consider images comprising of three channel components including Red, Green, and Blue (RGB). The value of each channel ranges from 0 to 255. In spite of the wide adoption by current implementations on shadow removal ~\cite{wang2018stacked,cun2020towards,zhu2022efficient}, the shadow regions often severely deteriorate if represented in the RGB style.
Consequently, another color space named LAB has aroused interests of many researchers ~\cite{hu2019direction,liu2021shadow_b,chen2021canet}, where L represents perceptual luminance, A stands for the two unique colors of human vision: red and green while B corresponds to blue and yellow.
Note that the value of the L channel changes from 0 to 100 while these of the A and B channels vary from -127 to 128.
The LAB color space is a device-independent color system based on physiological characteristics and models the visual perception of human eyes.

Such a representation method stores the shadow-related luminance information in the L channel. Changes in the L channel do not affect the color information since it is stored in the A and B channels, and vice versa. Therefore, LAB color space has been demonstrated to be more suitable for shadow removal.
\section{Methodology}

In this section, we first introduce the overall framework of the proposed LAB-Net. Then, we construct the Basic Block of the network through a parallel structure, which can aggregate more powerful features while remaining lightweight. Finally, we introduce a local spatial attention module that fully uses the prior information of non-shadow regions.

\subsection{LAB-Net Framework}

Motivated by the fact that shadow removal is highly correlated with image luminance, we choose to represent images in the LAB color space for its native advantages of separate luminance in the L channel and colors in the A \& B channels. In particular, we process the luminance and color channels in parallel and encourage them to mutually learn from each other for better information exchange. Therefore, we construct our LAB-Net by a two-branch structure shown in Fig.\,\ref{fig:LAB-Net}, which excavates both luminance and color information in the LAB color space.

Specifically, given a 3-channel shadow image $I_{\text{shadow}}$ in the LAB color space, associated with a 1-channel shadow mask $I_{\text{mask}}$, the shadow-free output $I_{\text{free}}$ of our LAB-Net is formulated as:
\begin{equation}
    I_{\text{free}} = G_{\text{LAB-Net}}(I_{\text{shadow}},
    I_{\text{mask}}),
\end{equation}
where $G_{\text{LAB-Net}}$ represents our proposed LAB-Net network. At the beginning of our LAB-Net's processing, we arrange a 3-channel 3$\times$3 convolution to process the shadow image $I_{\text{shadow}}$ first, whose result  is then concatenated with the shadow mask $I_{\text{mask}}$ to form a 4-channel primary feature as:

\vspace{-1em}
\begin{gather}
    I_{0} = Concat\big[Conv_3(I_{\text{shadow}}), I_{\text{mask}}\big].
\end{gather}

\begin{figure*}[!t]
    \centering
    \includegraphics[width=1.0\linewidth]{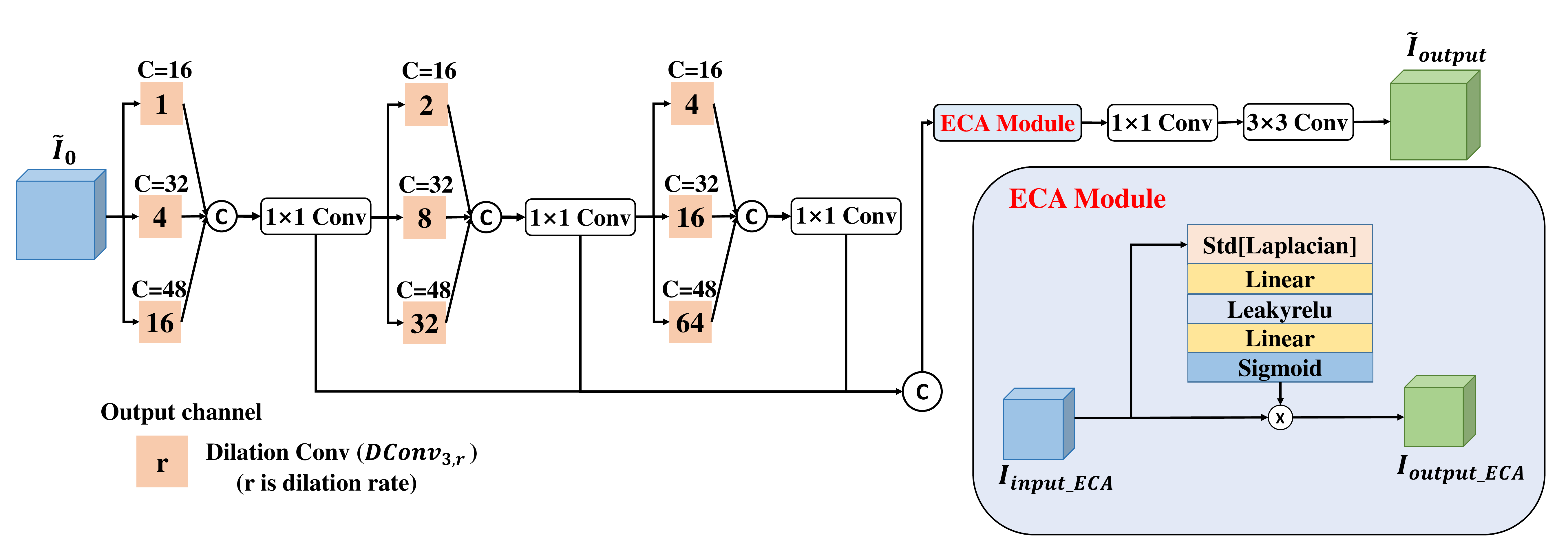}
    \caption{Basic Block in LAB-Net.}
    \label{fig:Basic block}
\end{figure*}
Then, the primary feature, $I_{0}$, is fed to the subsequent two-branch structure for processing the luminance and color information. Each branch mainly consists of several sequentially-stacked Basic Blocks $BB(\cdot)$, and a local spatial attention module (LSA) is inserted between each pair of Basic Blocks. Details of the Basic Block and LSA module are specified in the following two subsections.
The first branch processes the color information (A \& B channels) while the second branch processes the luminance information (L channel).
Specifically, the primary feature $I_0$ is respectively fed to the first Basic Blocks in the two branches:

\vspace{-1em}
\begin{gather}
    I_{1}^{\text{AB}} = BB^{\text{AB}}_{1}(I_0), \\
    I_{1}^{\text{L}} = BB^{\text{L}}_{1}(I_0).
\end{gather}

Then, $I_{1}^{\text{AB}}$ and $I_{1}^{\text{L}}$ are concatenated together. In each branch, a 32-channel 1$\times$1 convolution is applied to exchange luminance and color information, and the second Basic Block is adopted to refine the features:

\vspace{-1em}
\begin{gather}
    I_2^{\text{AB}} = BB^{\text{AB}}_2\Big((Conv_1)^{\text{AB}}_1\big(Concat[I^{\text{AB}}_{1},I^{\text{L}}_{1}]\big)\Big), \\
    I_2^{\text{L}} = BB^{\text{L}}_2\Big((Conv_1)^{\text{L}}_1\big(Concat[I^{\text{AB}}_{1},I^{\text{L}}_{1}]\big)\Big).
\end{gather}

Similarly, $I_2^{AB}$ and $I_2^L$ are concatenated and 1$\times$1 convolution is applied. Differently, an LSA module is inserted prior to applying the Basic Block, thus we have:

\vspace{-1em}
\begin{gather}
    I_3^{\text{AB}} = BB^{\text{AB}}_3\bigg(LSA\Big((Conv_1)^{\text{AB}}_2\big(Concat[I^{\text{AB}}_{2},I^{\text{L}}_{2}]\big)\Big)\bigg)\label{I3AB}, \\
    I_3^{\text{L}} = BB^{\text{L}}_3\bigg(LSA\Big((Conv_1)^{\text{L}}_2\big(Concat[I^{\text{AB}}_{2},I^{\text{L}}_{2}]\big)\Big)\bigg)\label{I3L}.   
\end{gather}

Note that, the $LSA$ module is applied here to use the information of non-shadow regions to assist the learning of shadow regions.
Then, we recursively apply Eq.\,(\ref{I3AB}) and Eq.\,(\ref{I3L}) based on $I_3^{\text{AB}}$ and $I_3^{\text{L}}$ to obtain a 2-channel feature $I_4^{\text{AB}}$ and 1-channel feature $I_4^{\text{L}}$. Finally, the original shadow image $I_{\text{shadow}}$ is added to the concatenation of $I_4^{\text{AB}}$ and $I_4^{\text{L}}$, whose result is convoluted with a 3-channel 3$\times$3 convolution to derive the eventual shadow-free image $I_{\text{free}}$:

\vspace{-1em}
\begin{equation}
    I_{\text{free}} = Conv_{3}\big(Concat[I^{\text{AB}}_4, I^{\text{L}}_4] + I_{\text{shadow}}\big).
\end{equation}

\subsection{Basic Block}

Shadow removal restores the shadow regions for concordant with the non-shadow regions. The quality of cleansed images is improved when injecting multi-scale context information due to the complex image contents and pixel correlations. 
To excavate such information, we build the Basic Blocks of our LAB-Net as illustrated in Fig.\,\ref{fig:Basic block}. Technically, a Basic Block consists of an elementary unit to absorb multi-scale contexts with different computational costs and an enhanced channel attention module to aggregate features at different levels.

\subsubsection{Elementary Unit}

Many previous studies have demonstrated the efficacy of dilated convolutions in shadow removal tasks. For example, DHAN~\cite{cun2020towards} constructs a series of dilated convolutions with linearly increasing dilation rates and is demonstrated to not only capture multi-scale information but also low-level features.

However, the dilated convolutions are arranged in a tandem fashion where low-level features are processed with a small dilation rate while a large one is applied to catch high-level features. As a result, both low-level and high-level features are not fully explored. Also, computational costs are large since the dilated convolutions with various dilation rates use an identical number of channels to process inputs.

As shown in Fig.\,\ref{fig:Basic block}, our elementary unit is a three-stage process by considering the above issues. Each stage is mainly comprised of three parallelized built-in dilated convolutions of divergent dilation rates to receive multi-scale contexts from the same input. Considering an input feature $\tilde{I}_0$, we feed it to the elementary unit, and the output of the $n$-th stage is:
\begin{figure}[!t]
    \centering
    \begin{subfigure}[ht]{0.22\linewidth}
        \includegraphics[width=\linewidth, height=\linewidth]{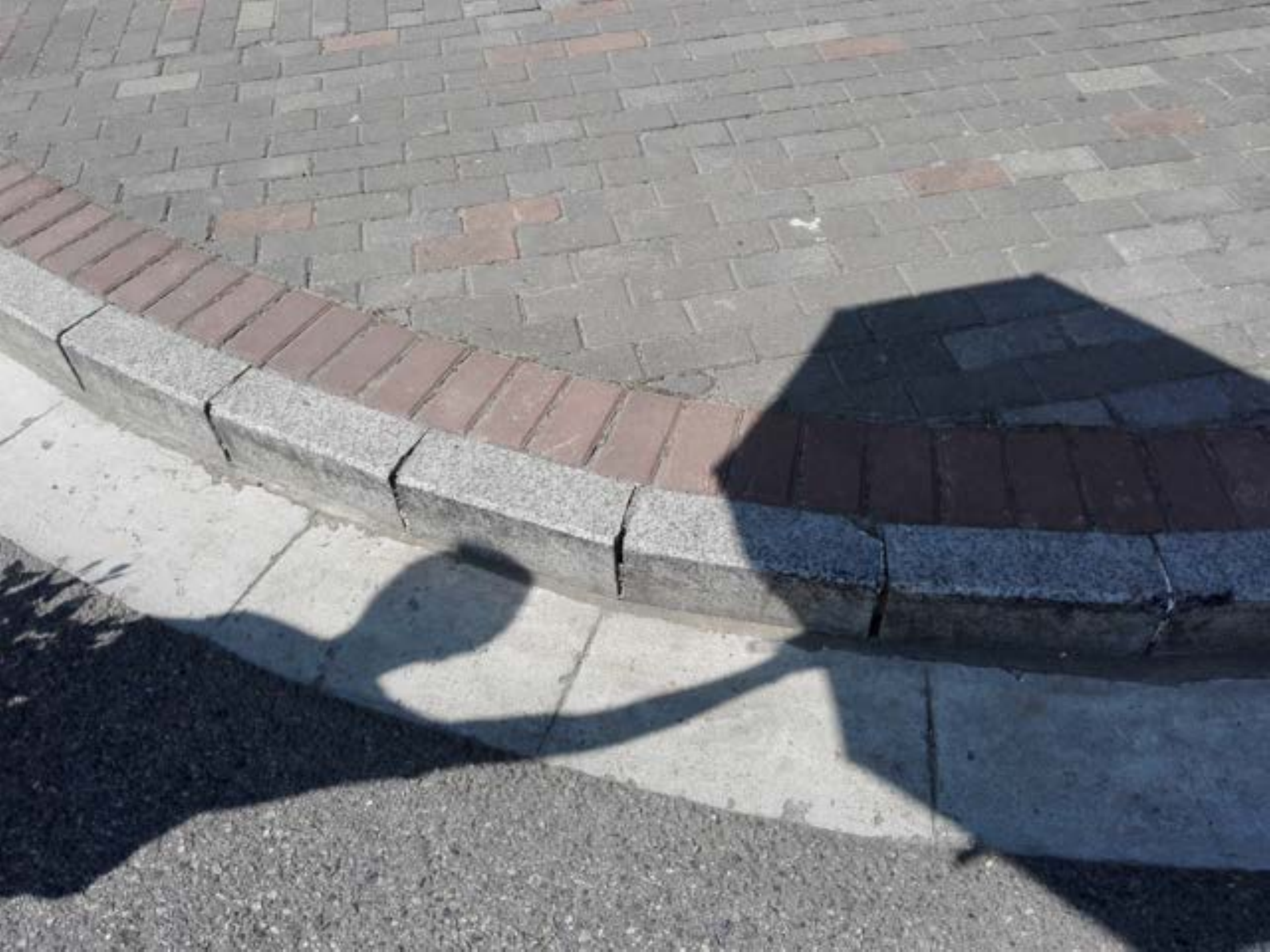}
    \caption{$I_{\text{shadow}}$}
   \end{subfigure}
    \begin{subfigure}[ht]{0.22\linewidth}
        \includegraphics[width=\linewidth, height=\linewidth]{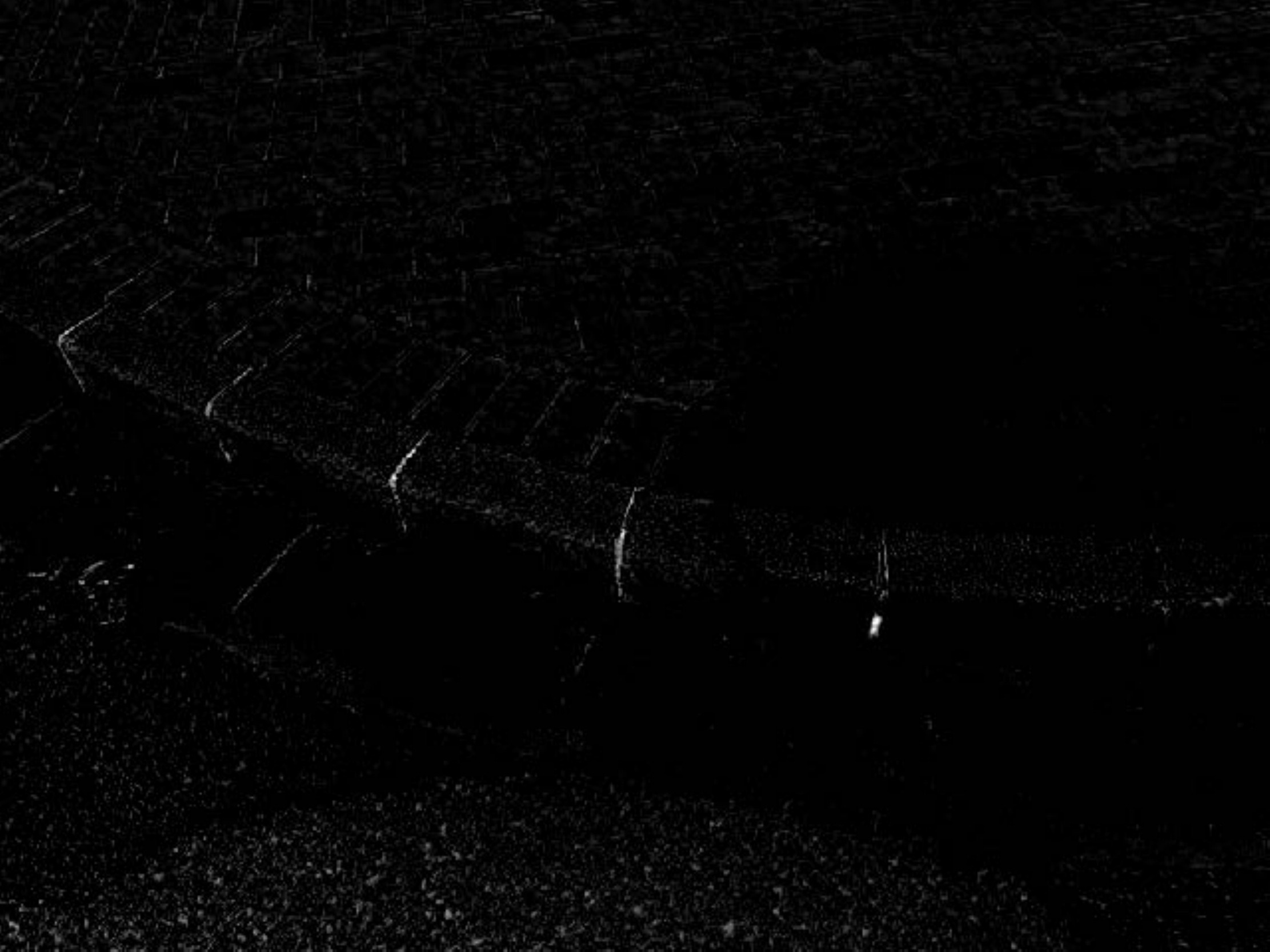}
    \caption{Feature}
   \end{subfigure}
    \begin{subfigure}[ht]{0.22\linewidth}
        \includegraphics[width=\linewidth, height=\linewidth]{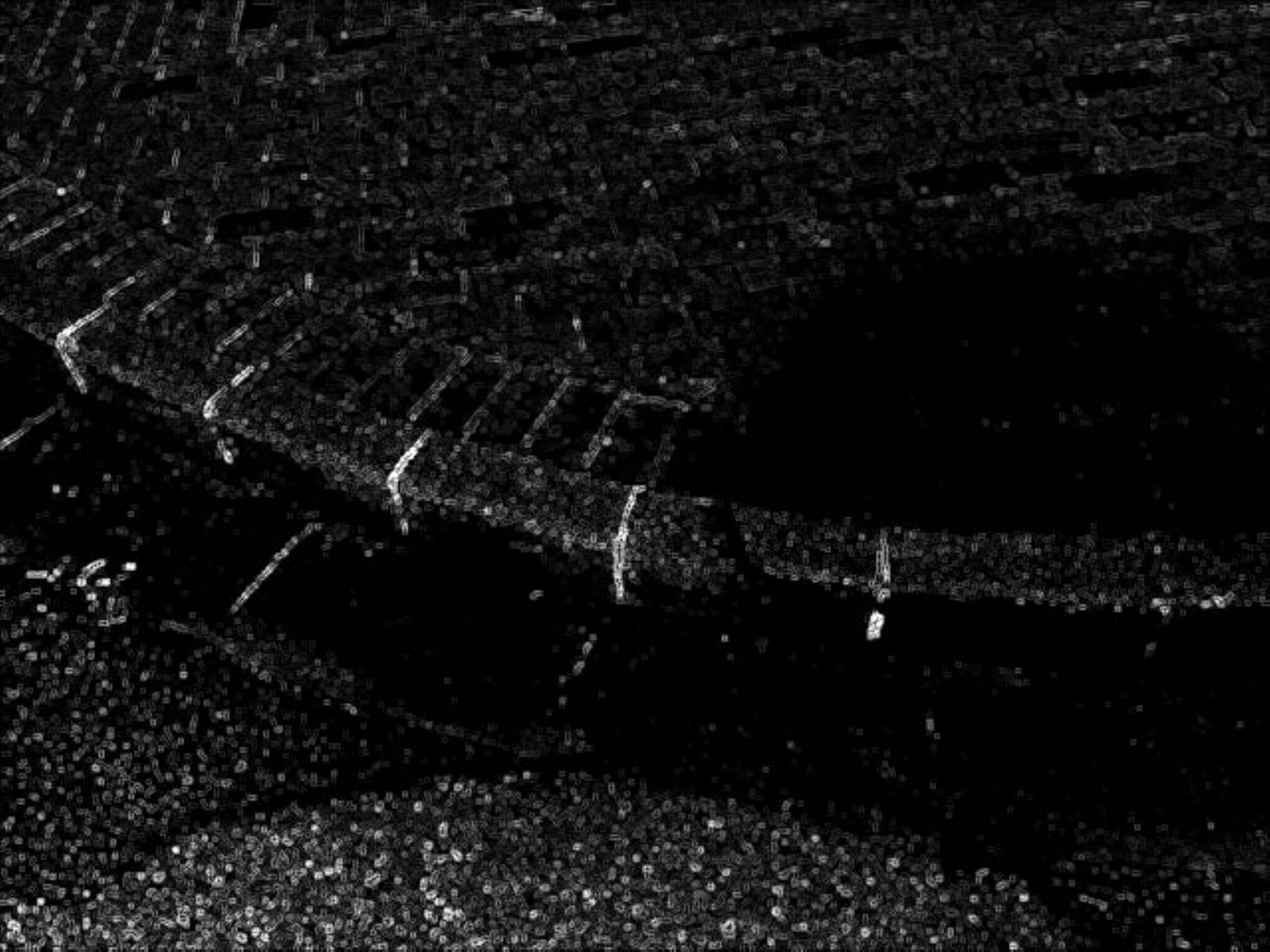}
    \caption{Sobel}
   \end{subfigure}
    \begin{subfigure}[ht]{0.22\linewidth}
        \includegraphics[width=\linewidth, height=\linewidth]{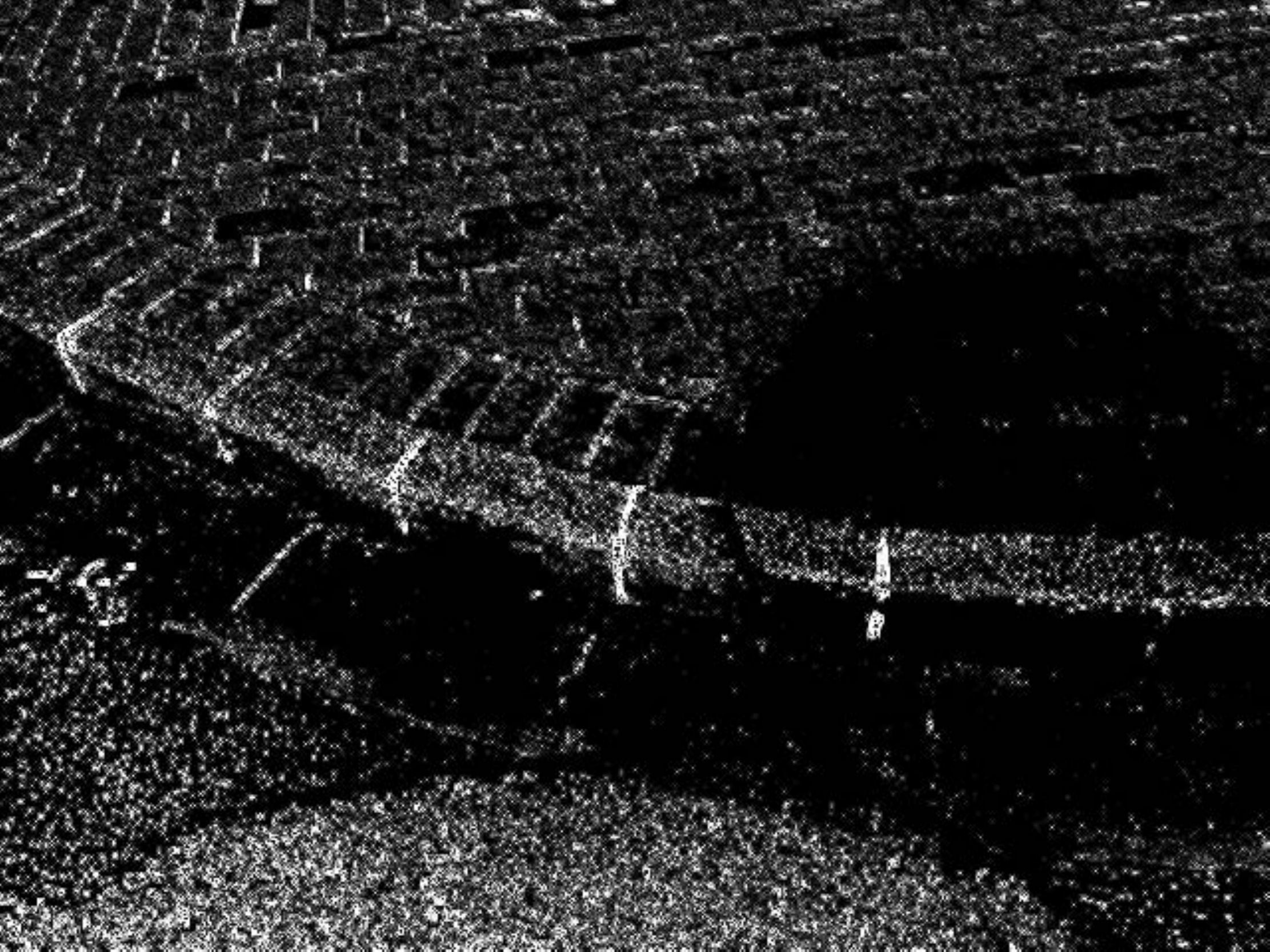}
    \caption{Laplacian}
   \end{subfigure}
    \caption{Visualization of shadow image, intermediate features of $I_2^L$ in Eq.\,(\ref{I3L}), Sobel filter and Laplacian filter.}
    \label{fig:Laplacian_vis}
\end{figure}

\begin{equation}
\begin{split}
   \tilde{I}_{n} = Conv_{1}\Big(Concat\big[DConv_{3, r_{n,1}}(\tilde{I}_{n-1}), \\ DConv_{3, r_{n,2}}(\tilde{I}_{n-1}), DConv_{3, r_{n,3}}(\tilde{I}_{n-1})\big]\Big).
\end{split}
\end{equation}
where $DConv_{3, r_{n,i}}(\cdot)$ represents the 3$\times$3 dilated convolution with dilation rate of $r_{n,i}$. In our LAB-Net, the dilation rates in the three stages are respectively set to: $r_{1, 1} =1, r_{1,2} = 4, r_{1,3} = 16; r_{2, 1} = 2, r_{2,2} = 8, r_{2,3} = 32; r_{3, 1} = 4, r_{3, 2} = 16, r_{3, 3} = 64$. We concatenate outputs in the three stages and a 32-channel 1$\times$1 convolution is applied to reduce channels.
%
The three stages consider contextual information at low, medium, and high levels. Moreover, the parallelized dilated convolutions with different dilated rates provide multi-scale information within the same level of features.

More global information can be obtained from a larger perceptual field; however, the gridding effect also leads to the loss of information continuity~\cite{wang2018understanding}. Therefore, it is natural to reserve more computational costs. We achieve this goal by reserving more channels for the dilated convolutions with a larger dilation rate. For feature extraction at the same level, the channel number is reserved as 16-32-48 according to the dilation rate. In this way, the overall network complexity is significantly reduced. The quantitative results of the channel reservation strategy are presented in the experimentation section.

Finally, we further concatenate the low-, medium- and high-level features as the intermediate features, which serve as the input of our enhanced channel attention module to aggregate these features at different levels. The concatenated result is defined as:
\begin{equation}
I_{\text{input\_ECA}} = Concat[\tilde{I}_1, \tilde{I}_2, \tilde{I}_3].
\end{equation}

\subsubsection{Enhanced Channel Attention Module}
In order to aggregate features at different levels, we build an enhanced channel attention module (ECA) behind the elementary unit.
Specifically, we first characterize the spatial details in the input feature of $I_{\text{input\_ECA}}$ by calculating the standard deviation of the Laplacian-filtered pixels as:
\begin{equation}\label{Laplacian}\tilde{I}_{\text{input\_ECA}} = Std\big[Laplacian(I_{\text{input\_ECA}})\big].
\end{equation}
where $Std(\cdot)$ returns the standard deviation of its input and $Laplacian(\cdot)$ represents the Laplacian filter.
Eq.\,(\ref{Laplacian}) is mainly inspired by an earlier study~\cite{itu1999subjective} that modeled the spatial information for video quality assessment tasks by a Sobel filter.
Compared with the Sobel filter, we realize that the Laplacian filter is more suitable for the shadow removal task. As shown in Fig.\,\ref{fig:Laplacian_vis}, the Laplacian filter can better focus on details in the intermediate features, thereby providing more spatial information.

With more spatial information, we further enhance the input feature $I_{\text{input\_ECA}}$ in a squeeze-and-excitation (SE) manner~\cite{hu2018squeeze}, formulated as:

\vspace{-1em}
\begin{align}
  I_{\text{output\_ECA}} =I_{\text{input\_ECA}} \cdot \delta\big(W_2\cdot \eta(W_1 \cdot \tilde{I}_{\text{input\_ECA}})\big).
\end{align}
where $W_1$ and $W_2$ are weights of two fully-connected layers, $\delta(\cdot)$ and $\eta(\cdot)$ represent the sigmoid and leakyrelu functions respectively.

\subsection{Local Spatial Attention Module}
We further excavate the non-shadow regions to better process the shadow regions. To this end, we propose a local spatial attention module (LSA) that transfers valuable prior information from non-shadow to shadow regions.

\begin{figure}[!t]
    \centering
    \includegraphics[width=1.0\linewidth]{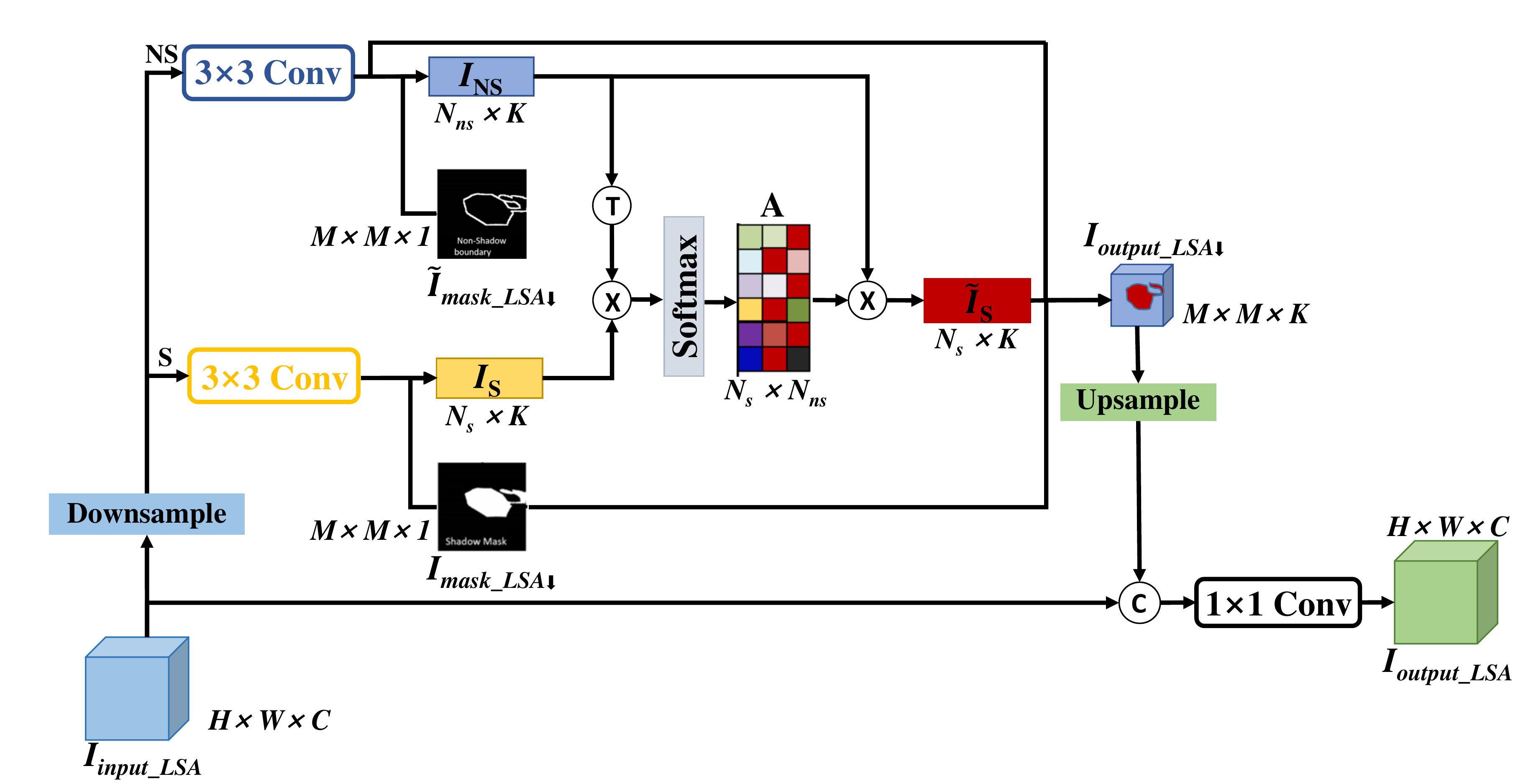}
    \caption{Structure of local spatial attention module (LSA).}
    \label{fig:LSA}
\end{figure}

Fig.\,\ref{fig:LSA} illustrates our LSA module. Note that, given an input $I_{\text{input}\_{\text{LSA}}} \in \mathbb{R}^{H \times W \times C}$ and its shadow mask $I_{\text{mask}\_{\text{LSA}}} \in \mathbb{R}^{H \times W \times 1}$, we first downsample them to $I_{\text{input}\_{\text{LSA}}\downarrow} \in \mathbb{R}^{M \times M \times C}$ and $I_{\text{mask}\_{\text{LSA}}\downarrow} \in \mathbb{R}^{M \times M \times 1}$, where $M \ll H$ and $M \ll W$ can reduce the computational complexity. 

Then, we pass $I_{\text{input\_LSA}\downarrow}$ to a K-channel 3$\times$3 convolution, together with the downsampled shadow mask $I_{\text{mask}\_{\text{LSA}}\downarrow}$ to obtain the shadow regions:
\vspace{-1em}
\begin{gather}
  I_{\text{S}} = Conv^{\text{S}}_{3}(I_{\text{input\_LSA} \downarrow})|I_{\text{mask}\_{\text{LSA}}\downarrow} = 1.
\end{gather}

Denoting $N_{s}$ as the number of shadow pixels, we rearrange $I_{\text{S}}$ with a shape of $\mathbb{R}^{N_{\text{s}} \times K}$. 

For non-shadow regions, in contrast to considering the whole non-shadow regions, we observe the local information in the boundary between shadow and non-shadow regions can be a useful way to avoid the influence of noise and irrelevant information, but also well reduce the computational cost of information match. To this end, we first dilate the shadow mask~\cite{haralick1987image} and the mask for non-shadow boundary is obtained by:  $\tilde{I}_{\text{mask}\_{\text{LSA}}\downarrow} = Dilation(I_{\text{mask}\_{\text{LSA}}\downarrow}) - I_{\text{mask}\_{\text{LSA}}\downarrow}$.
Thus, the non-shadow regions in the boundary can be obtained by:

\vspace{-1em}
\begin{gather}
  I_{\text{NS}} = Conv^{\text{NS}}_{3}(I_{\text{input\_LSA} \downarrow})| \tilde{I}_{\text{mask}\_{\text{LSA}} \downarrow} = 1.
\end{gather}

Denoting $N_{ns}$ as the number of non-shadow boundary pixels, we rearrange $I_{\text{NS}}$ with a shape of $\mathbb{R}^{N_{\text{ns}} \times K}$.

Inspired by the recent progress on the transformer network~\cite{vaswani2017attention}, we consider $I_{\text{S}}$ as a query matrix to calculate the spatial attention \emph{w.r.t.} $I_{\text{NS}}$ as the key matrix. Specifically, the attention is derived as:

\vspace{-1em}
\begin{align}
A = \sigma\big(I_{\text{S}} \cdot (I_{\text{NS}})^T\big).
\end{align}
where $\sigma(\cdot)$ represents the row-wise softmax operation. Then, the attention-transformed shadow regions becomes:

\vspace{-1em}
\begin{align}
\tilde{I}_{\text{S}} = A \cdot I_{\text{NS}}.
\end{align}

We copy $\tilde{I}_{\text{S}}$ and paste it to the shadow regions in $Conv^{\text{NS}}_{3}(I_{\text{input\_LSA}\downarrow})$, the result forms the output of our LSA module, denoted as $I_{\text{output\_LSA}\downarrow}$. Then, we upsample $I_{\text{output\_LSA}\downarrow}$ back to the shape of $\mathbb{R}^{H \times W \times K}$, denoted as $I_{\text{output\_LSA}\uparrow}$. Finally, we concatenate the input $I_{\text{input\_LSA}}$ with $I_{\text{output\_LSA}\uparrow}$ and pass through a C-channel 1$\times$1 convolution to obtain final outputs of the entire LSA module:

\vspace{-1em}
\begin{align}
I_{\text{output\_LSA}} = Conv_{1}\big(Concat[I_{\text{input\_LSA}},I_{\text{output\_LSA$\uparrow$}}]\big).
\end{align}

\begin{table*}[!t]
\caption{Performance comparison on ISTD dataset.The best and the second results are boldfaced and
underlined, respectively.}
\centering
\small
\setlength{\tabcolsep}{2.3pt}
\begin{tabular}{c|c|c|ccc|ccc|ccc}
\hline
\multicolumn{1}{c|}{\multirow{2}{*}{Methods}} & 
\multicolumn{1}{c|}{\multirow{1}{*}{Param}} &
\multicolumn{1}{c|}{\multirow{1}{*}{Flops}} &
\multicolumn{3}{c|}{Shadow Region (S)} & 
\multicolumn{3}{c|}{Non-Shadow Region (NS)} & 
\multicolumn{3}{c}{All image (ALL)} \\ \cline{4-12}
\multicolumn{1}{c|}{} & \multicolumn{1}{c|}{($10^6$)} &\multicolumn{1}{c|}{($10^9$)} & RMSE $\downarrow$ & PSNR $\uparrow$ & SSIM $\uparrow$ & RMSE $\downarrow$ & PSNR $\uparrow$ & SSIM $\uparrow$ & RMSE $\downarrow$ & PSNR $\uparrow$ & SSIM $\uparrow$ \\ \hline
\multicolumn{1}{c|}{Input Image}& \multicolumn{1}{c|}{-} &\multicolumn{1}{c|}{-} &32.10  &22.40  &0.9361  &7.09  &27.32  &0.9755  &10.88  &20.56  &0.8934  \\ \hline
\multicolumn{1}{c|}{Guo~\emph{et al.}} 
&\multicolumn{1}{c|}{-} &\multicolumn{1}{c|}{-} &17.44  &29.19  &0.9547  &15.13  &21.69  &0.7538  &15.48  &20.51  &0.6997  \\
\multicolumn{1}{c|}{ST-CGAN} 
&\multicolumn{1}{c|}{29.24} &\multicolumn{1}{c|}{17.88} &14.09  &32.60  &0.9604  &12.87  &25.01  &0.7580  &13.06  &23.95  &0.7080  \\
\multicolumn{1}{c|}{DSC}
&\multicolumn{1}{c|}{22.30} &\multicolumn{1}{c|}{123.47} &8.72  &34.65  &0.9835  &5.04  &31.26  &0.9691  &5.59  &29.00  &0.9438  \\
\multicolumn{1}{c|}{SP+M-Net} 
&\multicolumn{1}{c|}{141.18} &\multicolumn{1}{c|}{160.10} &9.64  &32.89  &0.9861  &7.73  &26.11  &0.9650  &7.96  &25.01  &0.9483  \\
\multicolumn{1}{c|}{DHAN} 
&\multicolumn{1}{c|}{21.75}  &\multicolumn{1}{c|}{262.87} &\underline{7.73} &35.53  &\underline{0.9882}  &5.29  &31.05  &0.9705  &5.66  &29.11  &0.9543  \\
\multicolumn{1}{c|}{LG-ShadowNet}
&\multicolumn{1}{c|}{5.70} &\multicolumn{1}{c|}{28.64} &10.72  &31.53  &0.9788  &5.34 &29.47  &0.9673  &6.08  &26.62  &0.9356  \\ 
\multicolumn{1}{c|}{Auto-Exposure} 
&\multicolumn{1}{c|}{143.01} &\multicolumn{1}{c|}{160.32} &7.91  &34.71  &0.9752  &5.51  &28.61  &0.8799  &5.89  &27.19  &0.8456  \\
\multicolumn{1}{c|}{G2R}
&\multicolumn{1}{c|}{22.76} &\multicolumn{1}{c|}{113.87} &10.72  &31.63  &0.9746  &7.55 &26.19  &0.9671  &7.85  &24.72  &0.9324  \\
\multicolumn{1}{c|}{DC-ShadowNet}
&\multicolumn{1}{c|}{21.16} &\multicolumn{1}{c|}{105.00} &11.43  &31.69  &0.9760  &5.81 &28.99  &0.9580  &6.57  &26.38  &0.9220  \\
\multicolumn{1}{c|}{EMDN} 
&\multicolumn{1}{c|}{10.05} &\multicolumn{1}{c|}{56.13} &8.29  &\underline{36.95}  &0.9867  &\underline{4.55}  &\underline{31.54}  &\textbf{0.9779}  &\underline{5.09}  &\underline{29.85}  &\textbf{0.9598}  \\ \hline
\multicolumn{1}{c|}{Ours} &\multicolumn{1}{c|}{0.93} &\multicolumn{1}{c|}{59.62} & \textbf{6.65}  & \textbf{37.17} &\textbf{0.9887}  &\textbf{4.49}  &\textbf{32.42}  &\underline{0.9727}  & \textbf{4.84}  & \textbf{30.49}  & \underline{0.9563}  \\ \hline
\end{tabular}
\label{Table:Quant_res_istd}
\normalsize
\end{table*}

\begin{figure*}[htbp]
\centering

\begin{subfigure}[t]{0.09\textwidth}
\includegraphics[width=\linewidth, height=\linewidth]{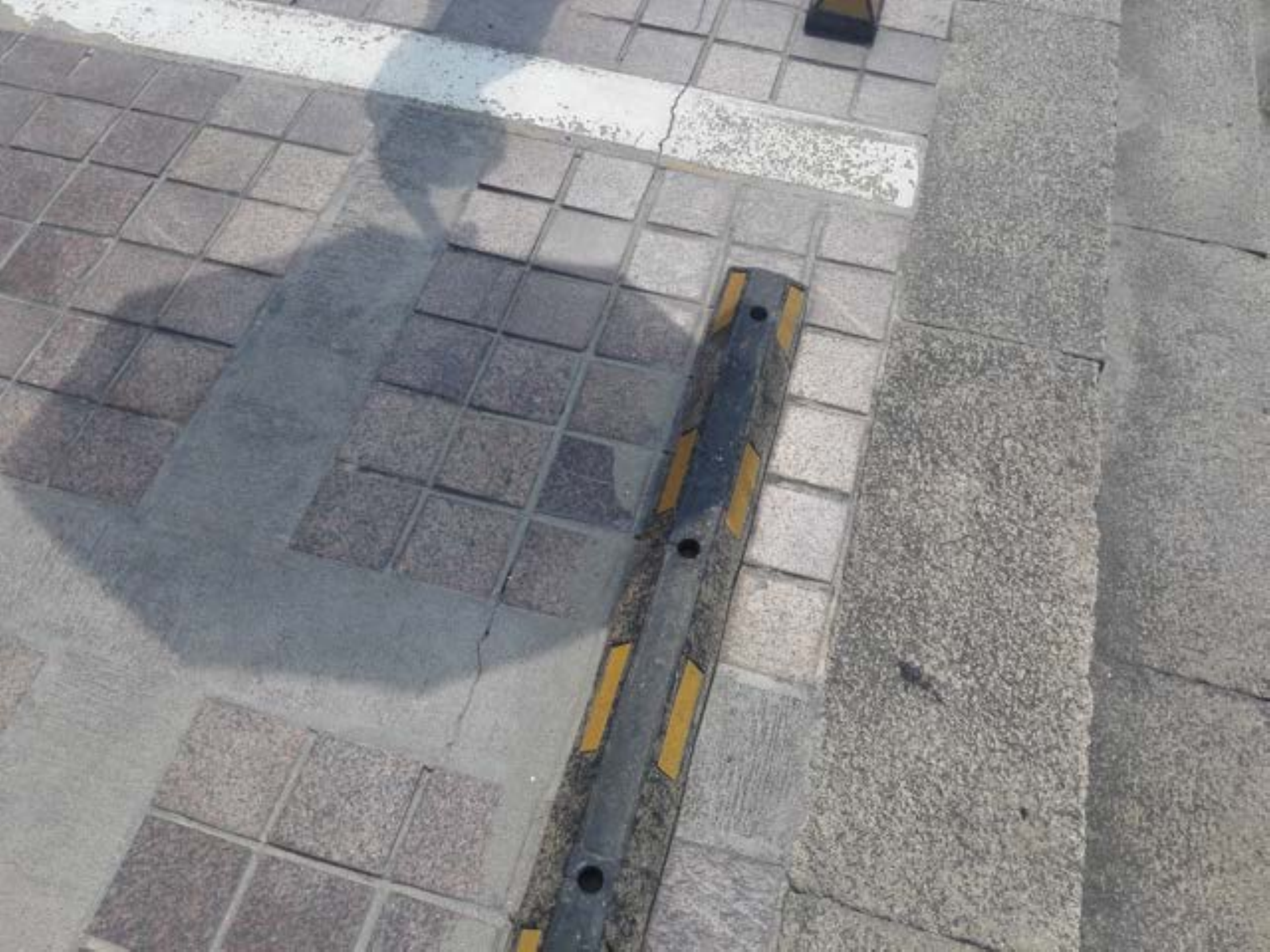} \\
\includegraphics[width=\linewidth, height=\linewidth]{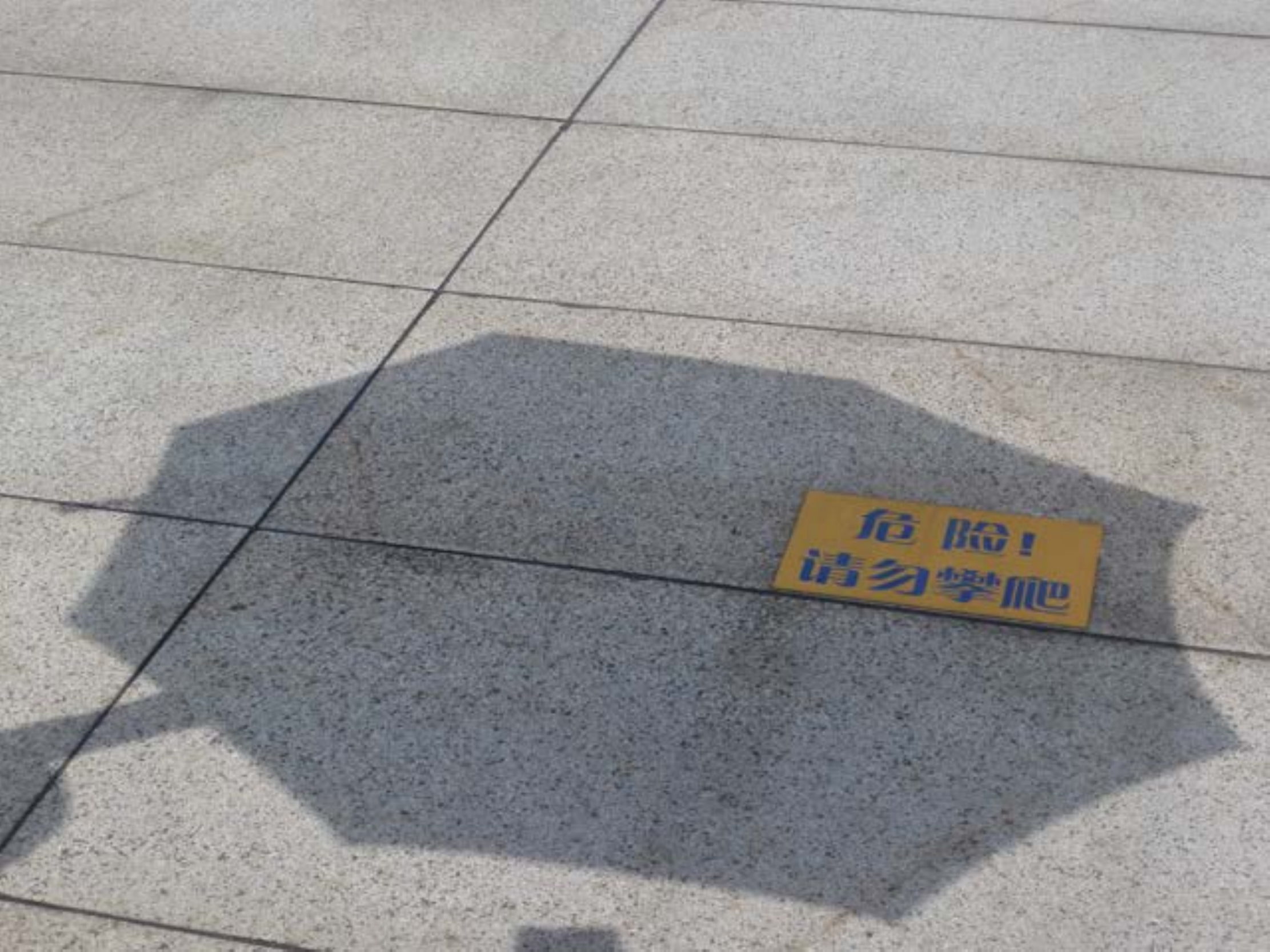} \\
\includegraphics[width=\linewidth, height=\linewidth]{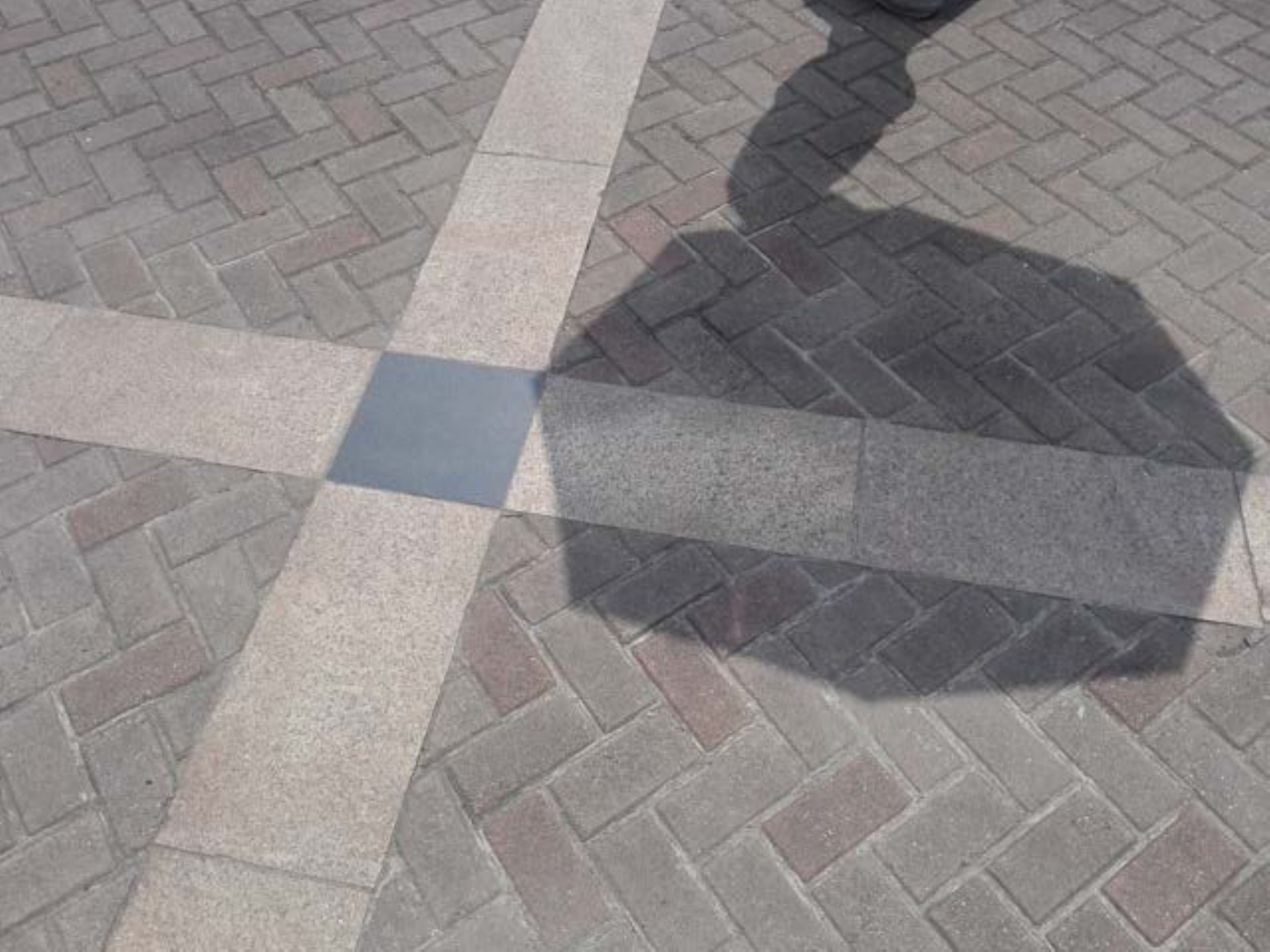} 
\caption{Input}
\end{subfigure}
\begin{subfigure}[t]{0.09\textwidth}
\includegraphics[width=\linewidth, height=\linewidth]{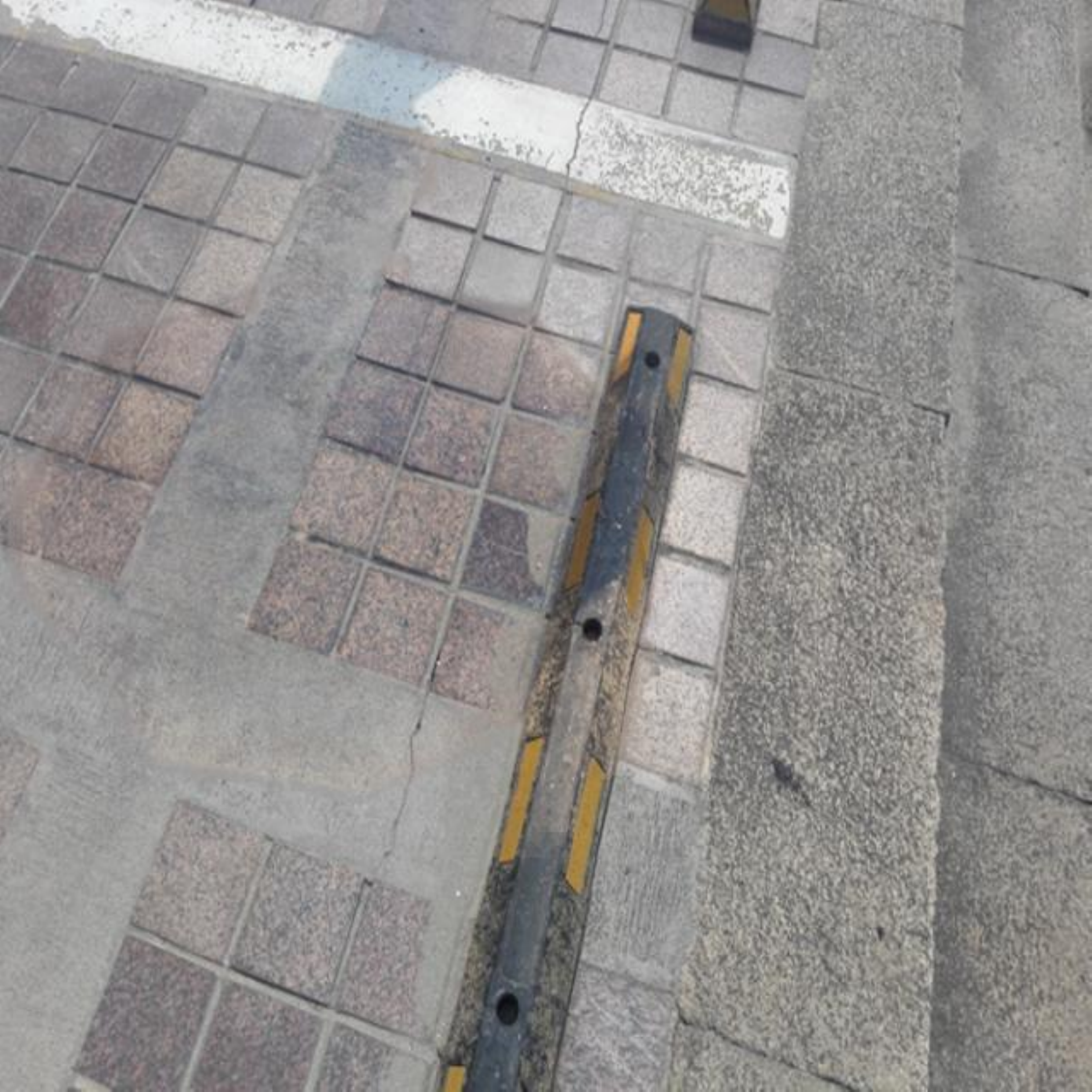} \\
\includegraphics[width=\linewidth, height=\linewidth]{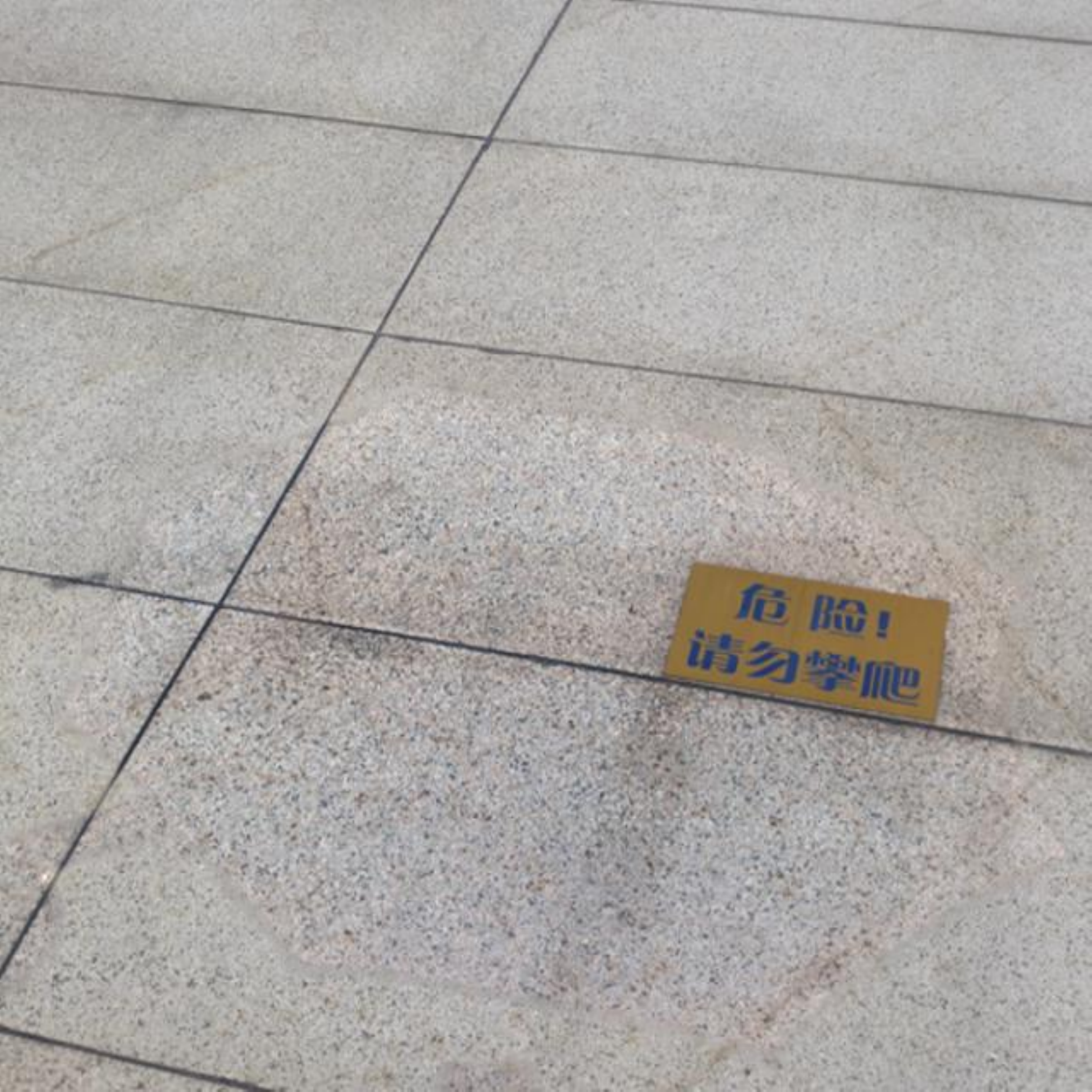} \\
\includegraphics[width=\linewidth, height=\linewidth]{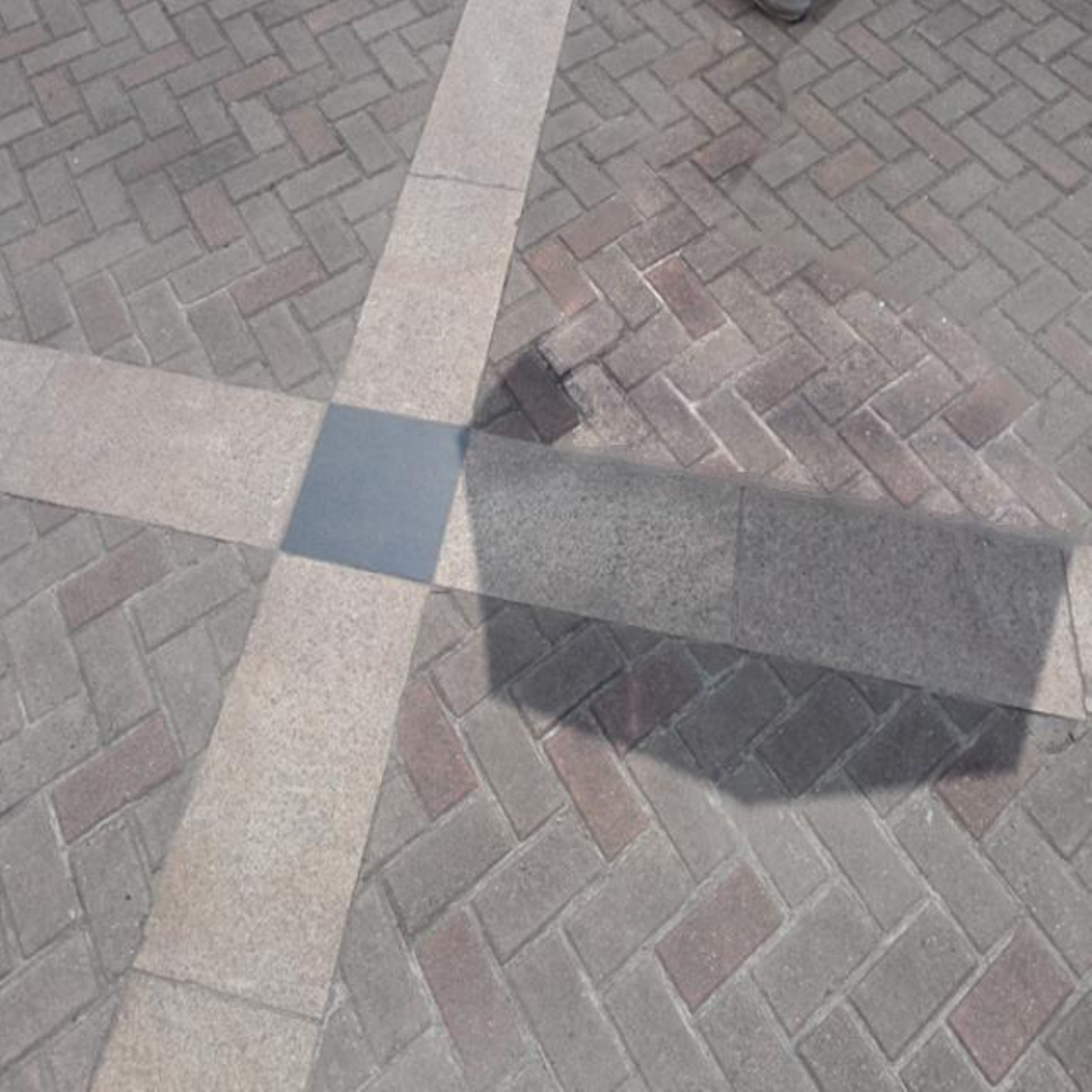}
\caption{Guo \emph{et al.}}
\end{subfigure}
\begin{subfigure}[t]{0.09\textwidth}
\includegraphics[width=\linewidth, height=\linewidth]{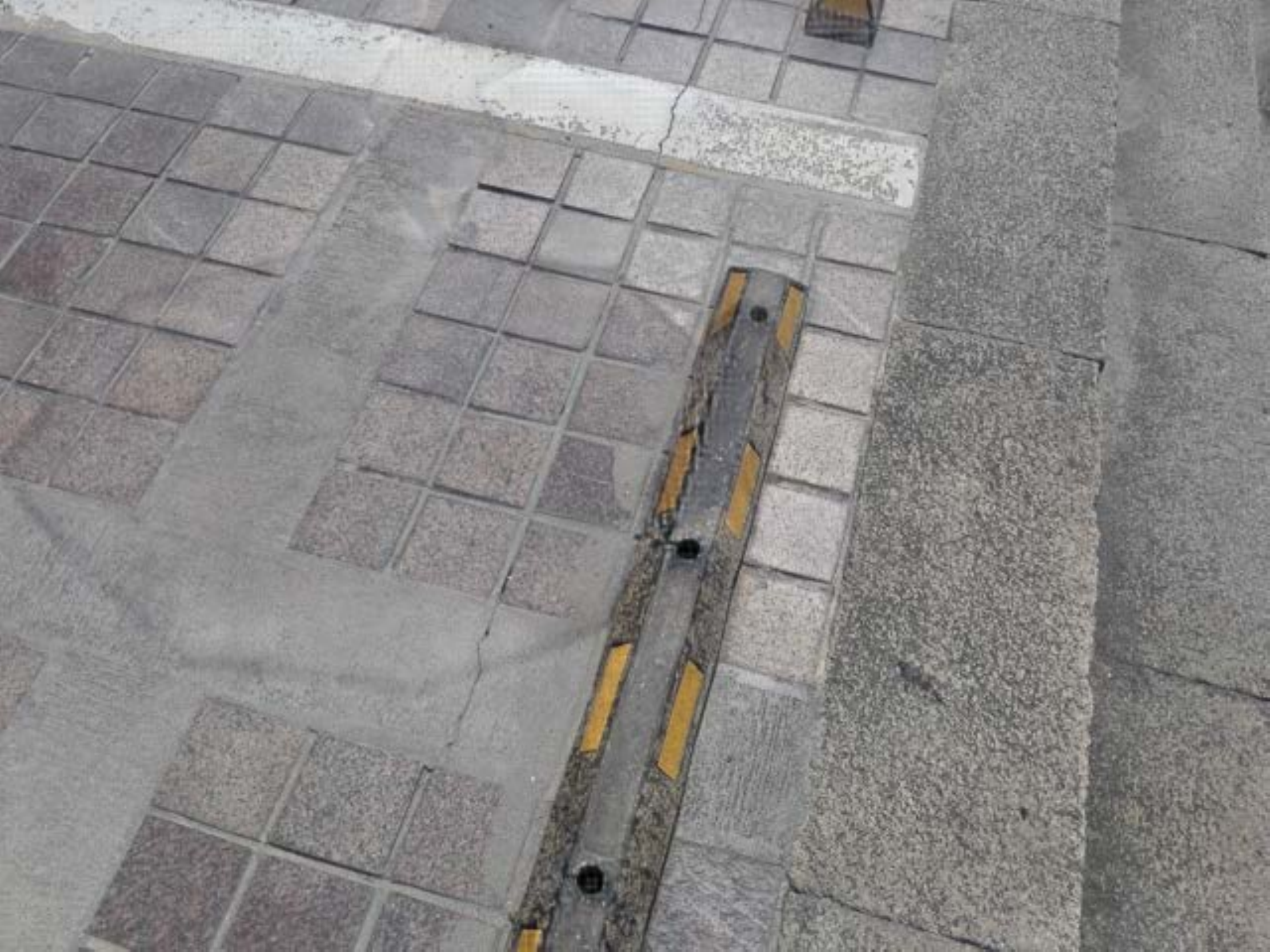} \\
\includegraphics[width=\linewidth, height=\linewidth]{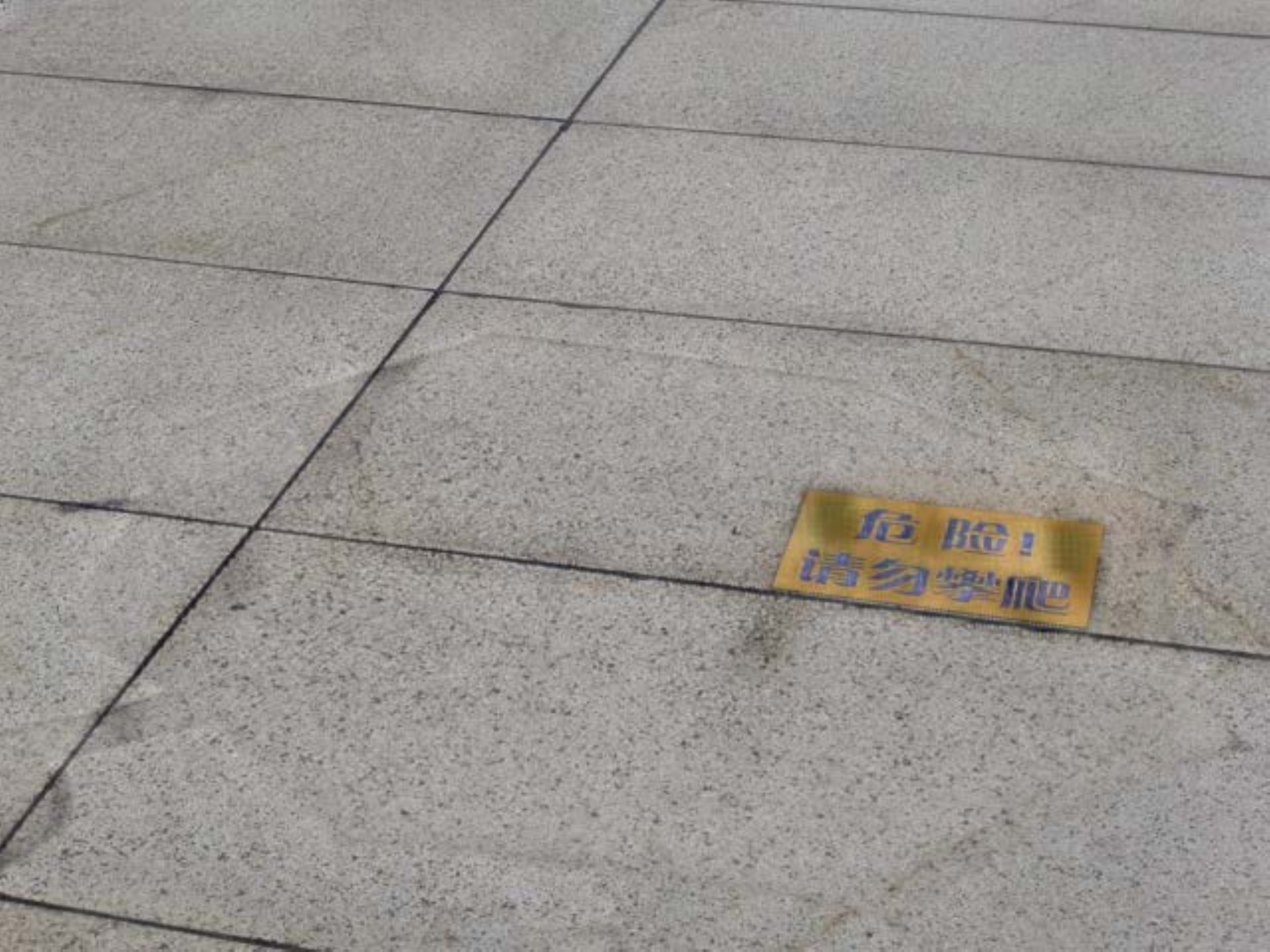} \\
\includegraphics[width=\linewidth, height=\linewidth]{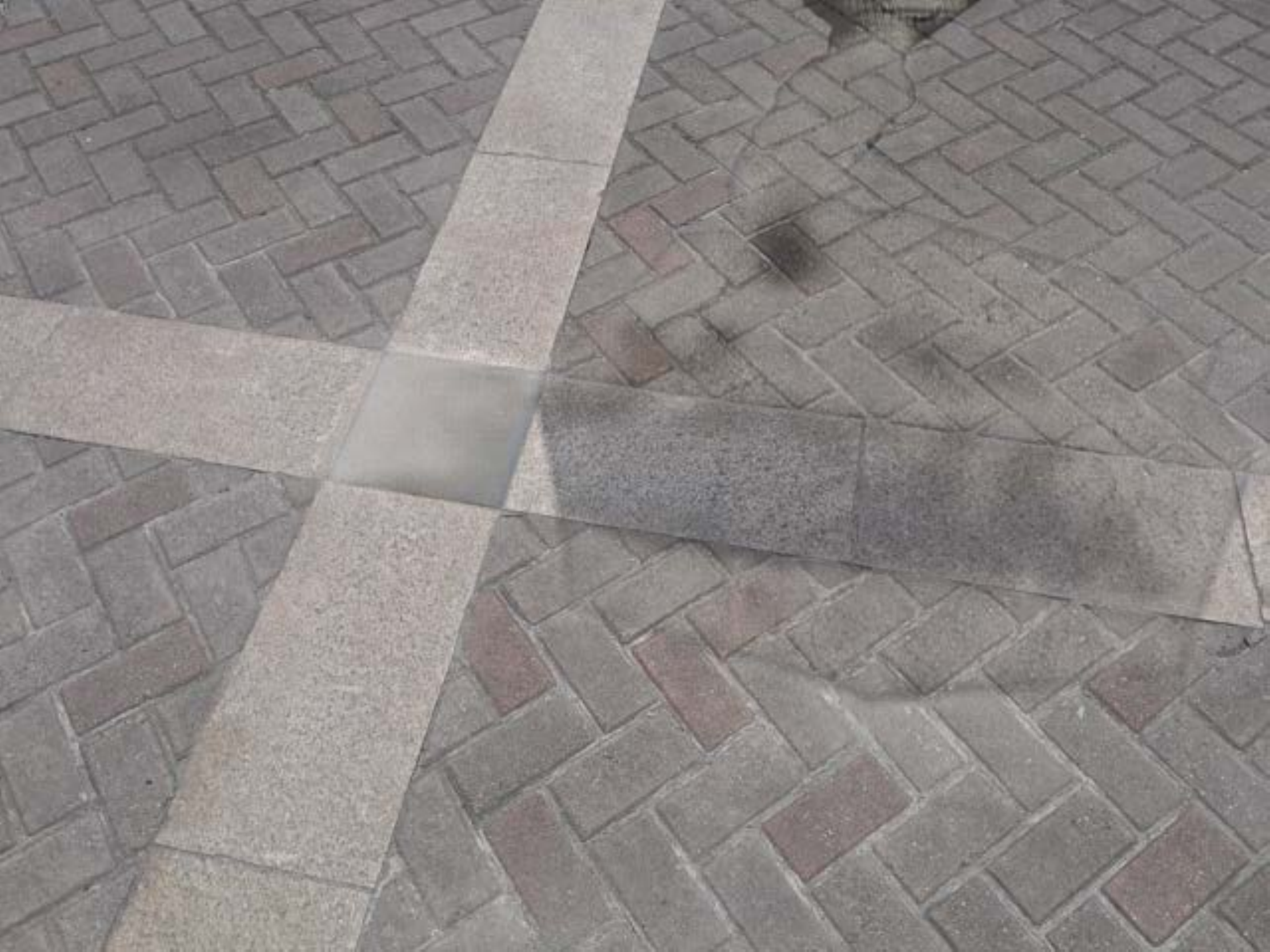} 
\caption{LG-SNet}
\end{subfigure}
\begin{subfigure}[t]{0.09\textwidth}
\includegraphics[width=\linewidth, height=\linewidth]{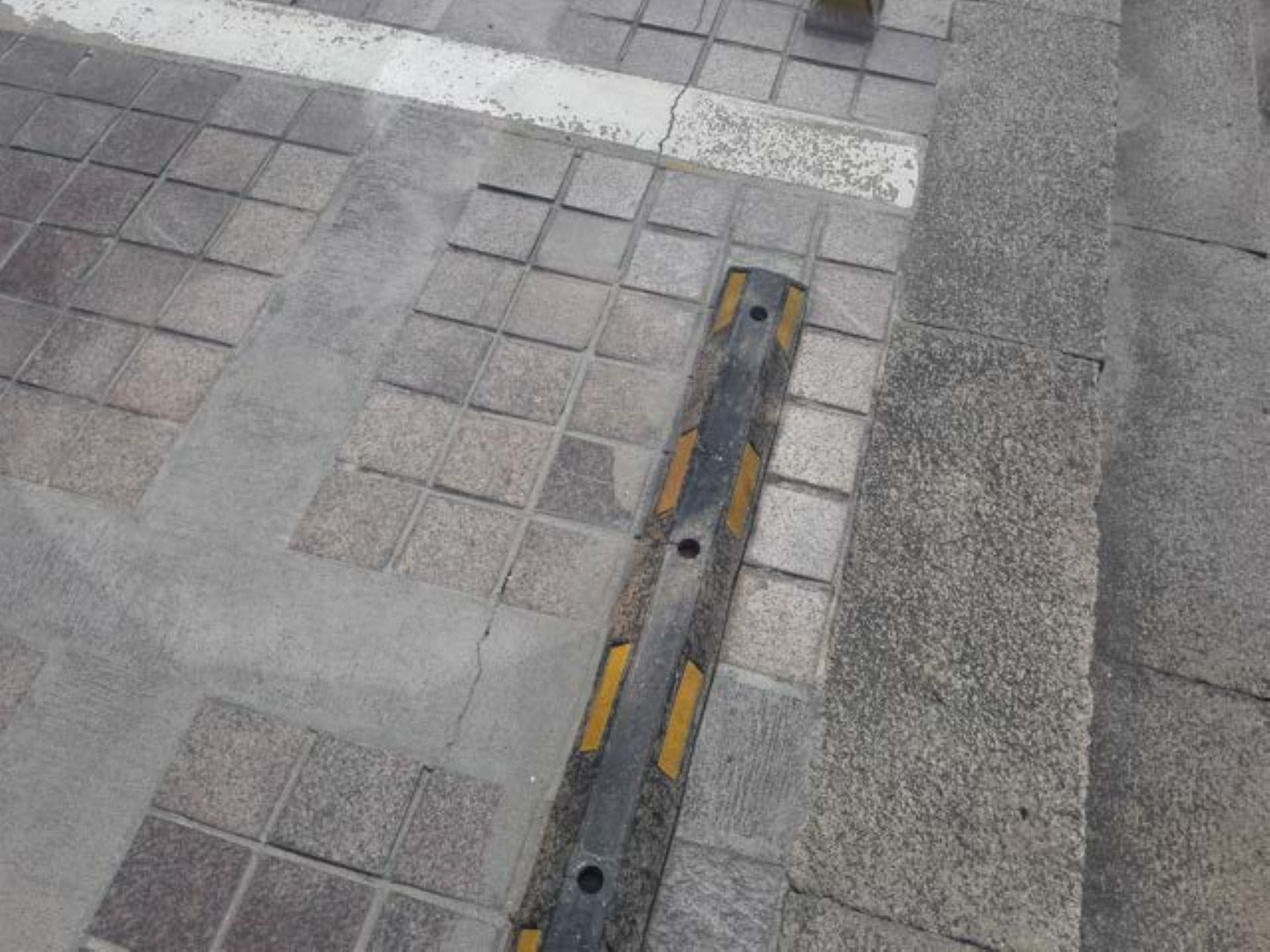} \\
\includegraphics[width=\linewidth, height=\linewidth]{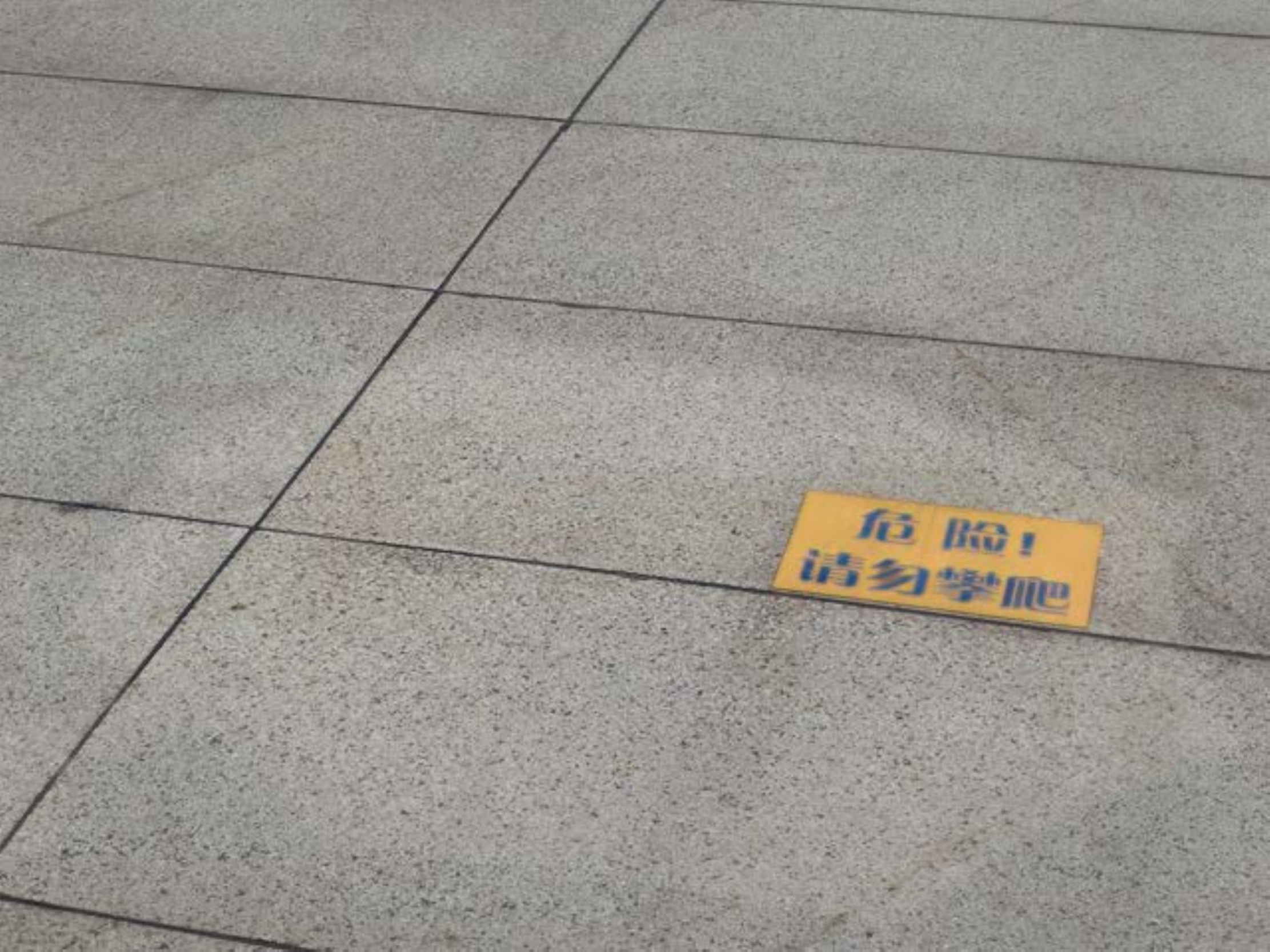} \\
\includegraphics[width=\linewidth, height=\linewidth]{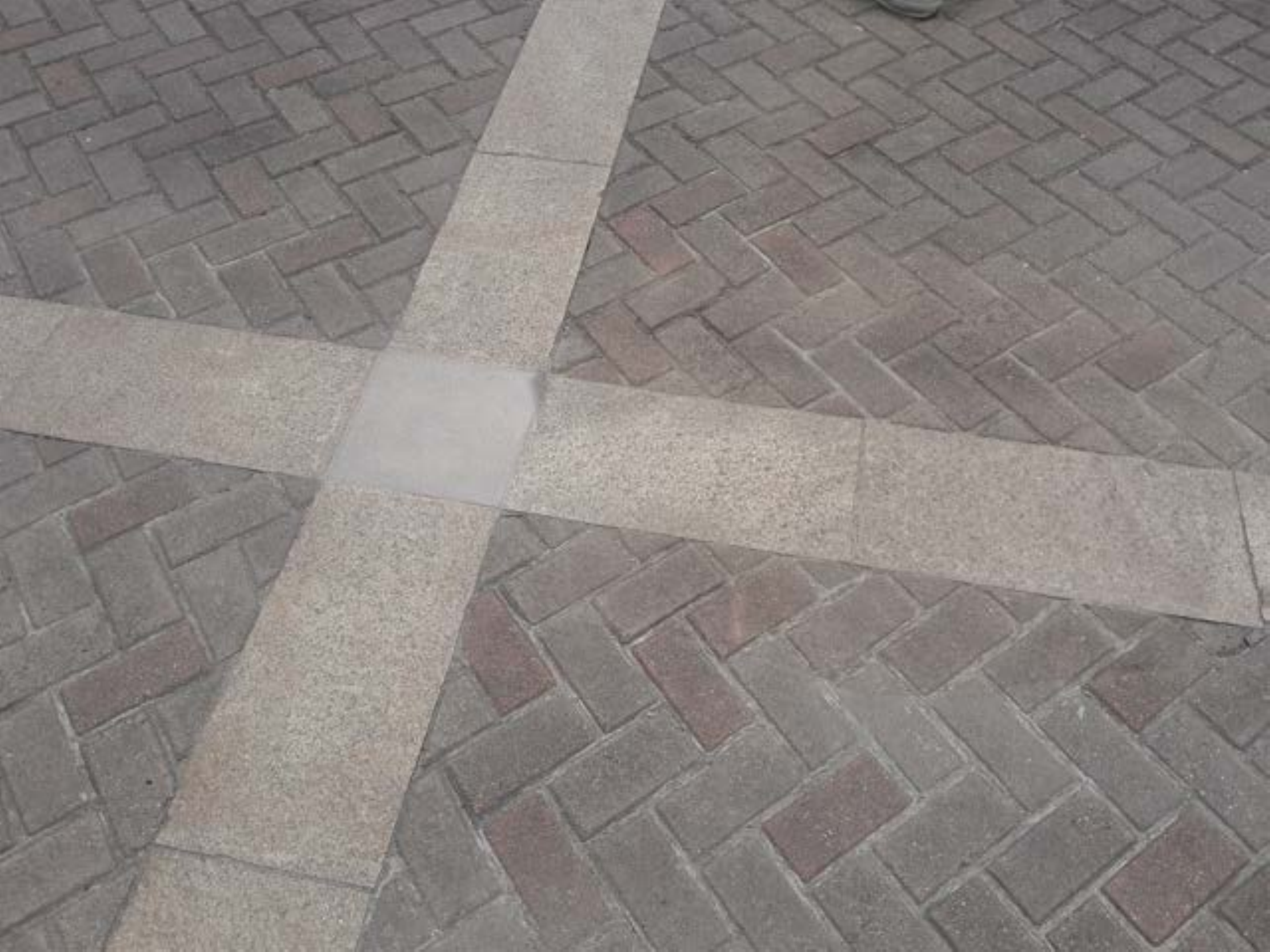} 
\caption{DHAN}
\end{subfigure}
\begin{subfigure}[t]{0.09\textwidth}
\includegraphics[width=\linewidth, height=\linewidth]{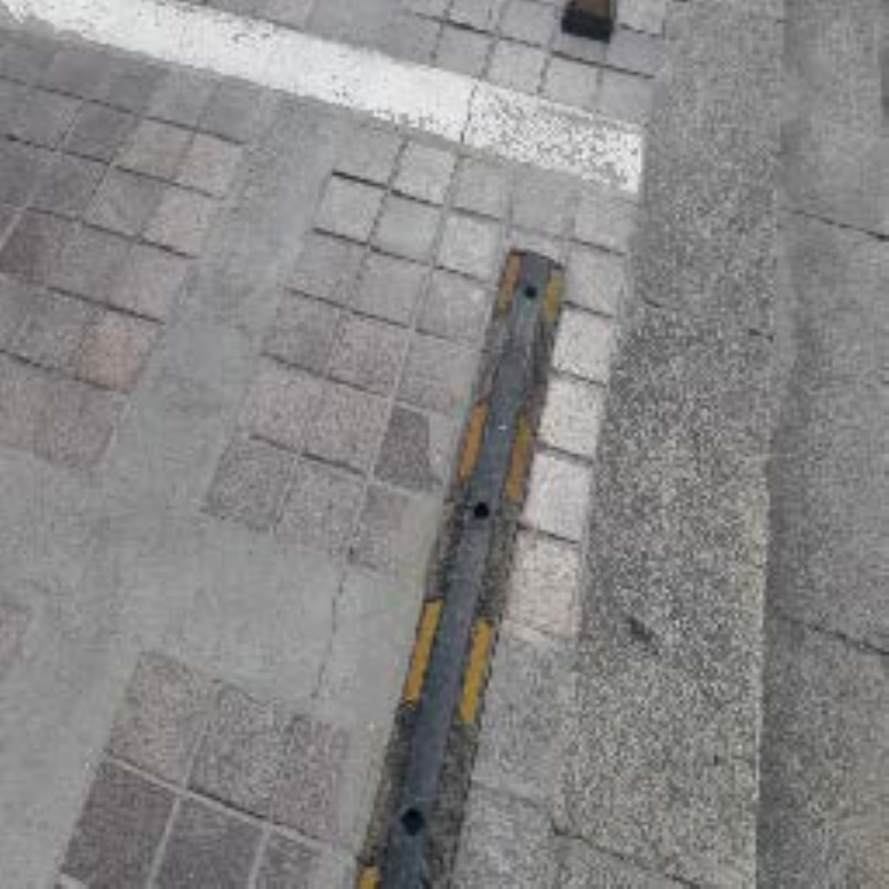} \\
\includegraphics[width=\linewidth, height=\linewidth]{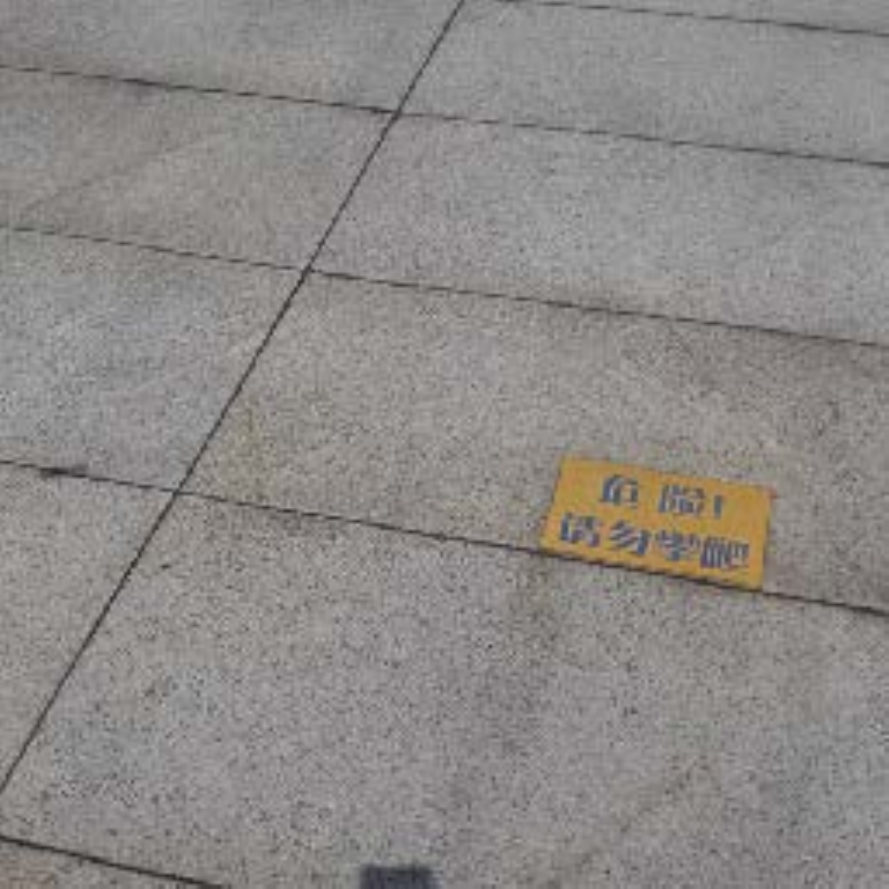} \\
\includegraphics[width=\linewidth, height=\linewidth]{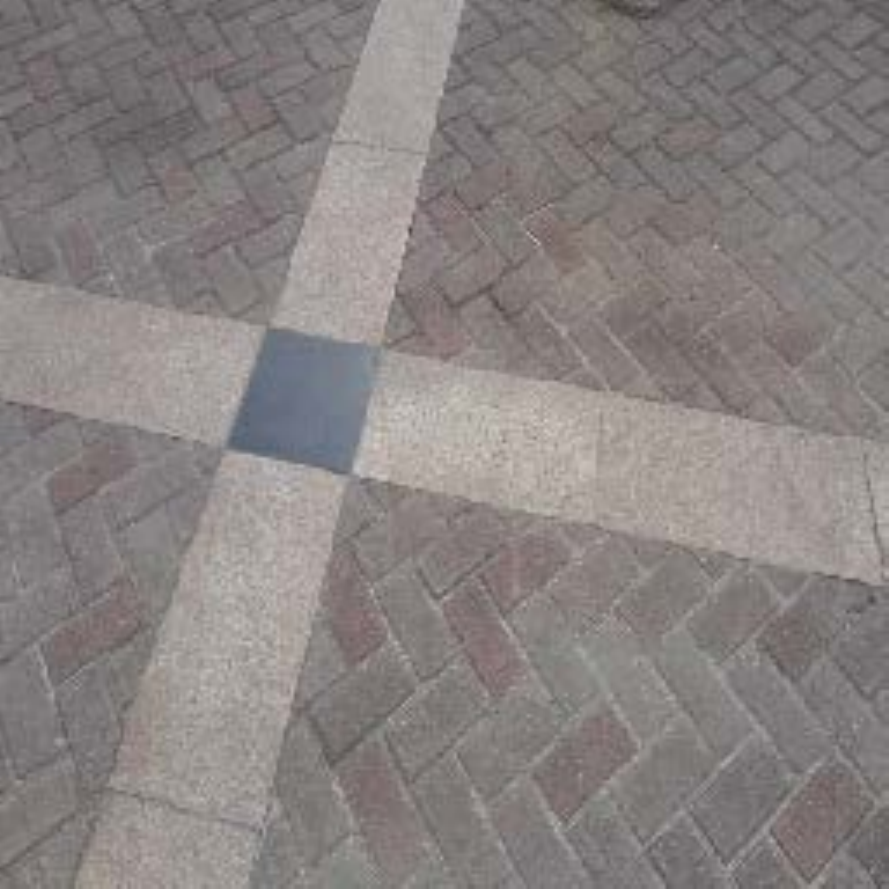}
\caption{A-E}
\end{subfigure}
\begin{subfigure}[t]{0.09\textwidth}
\includegraphics[width=\linewidth, height=\linewidth]{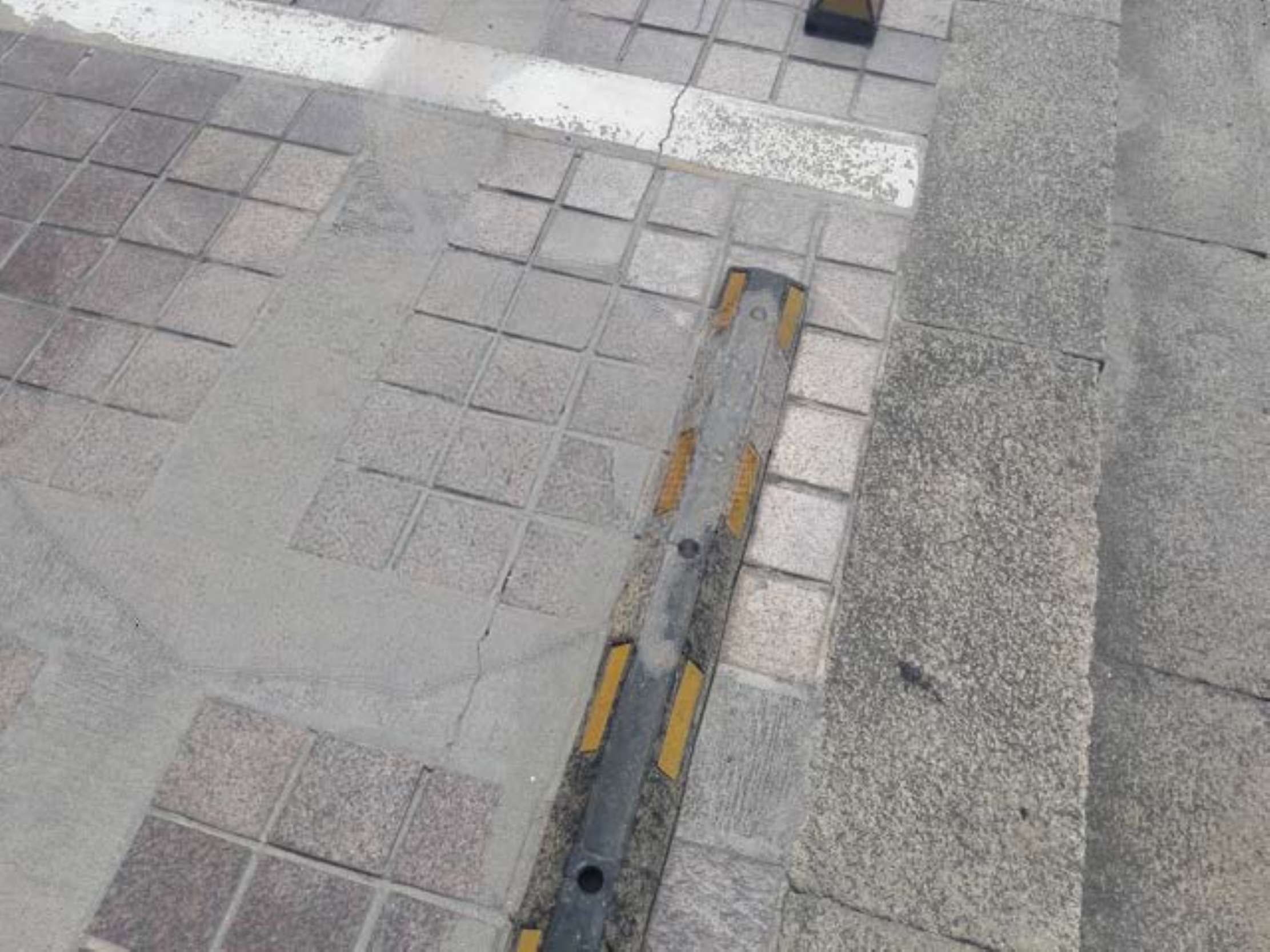} \\
\includegraphics[width=\linewidth, height=\linewidth]{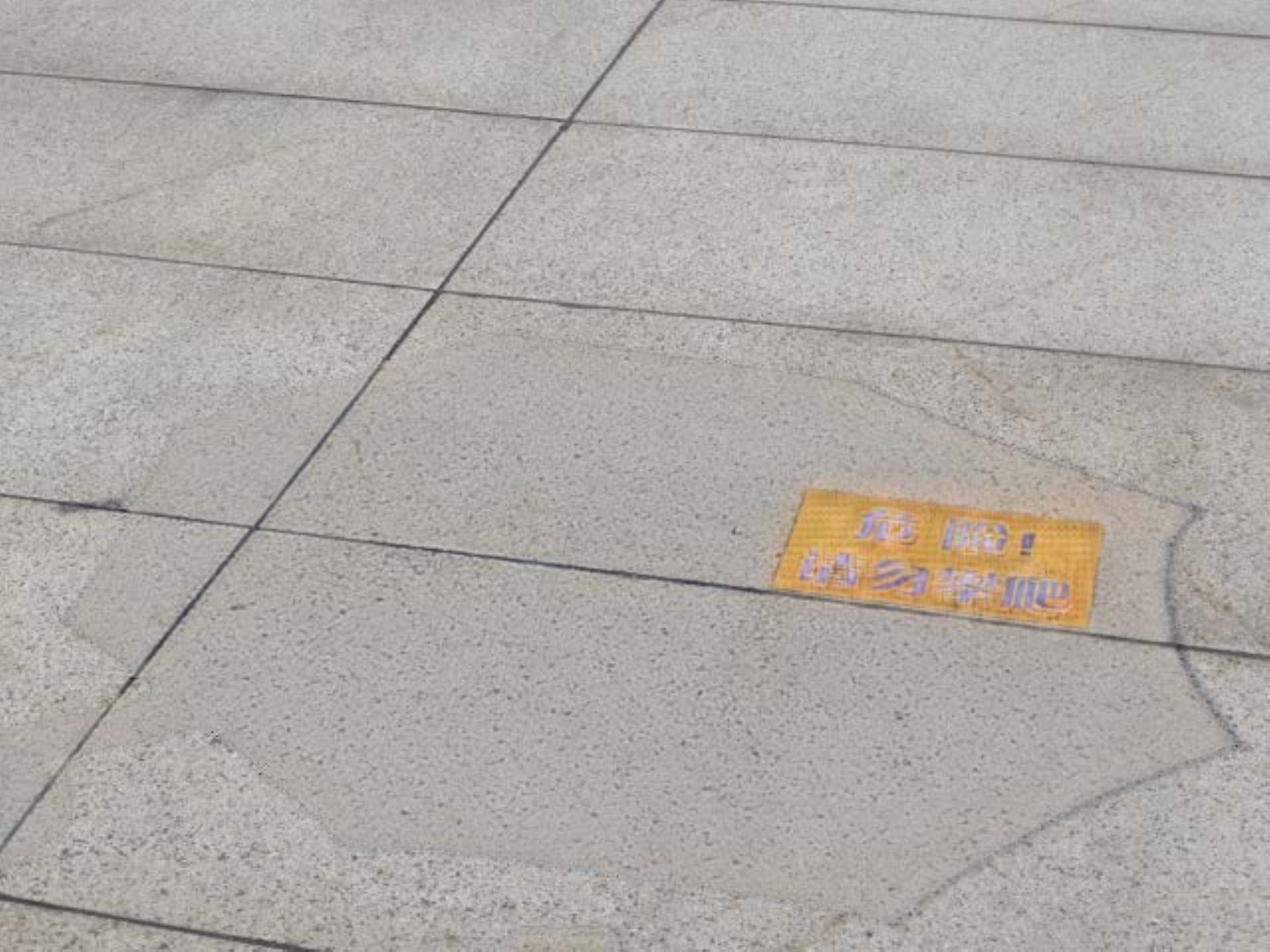} \\
\includegraphics[width=\linewidth, height=\linewidth]{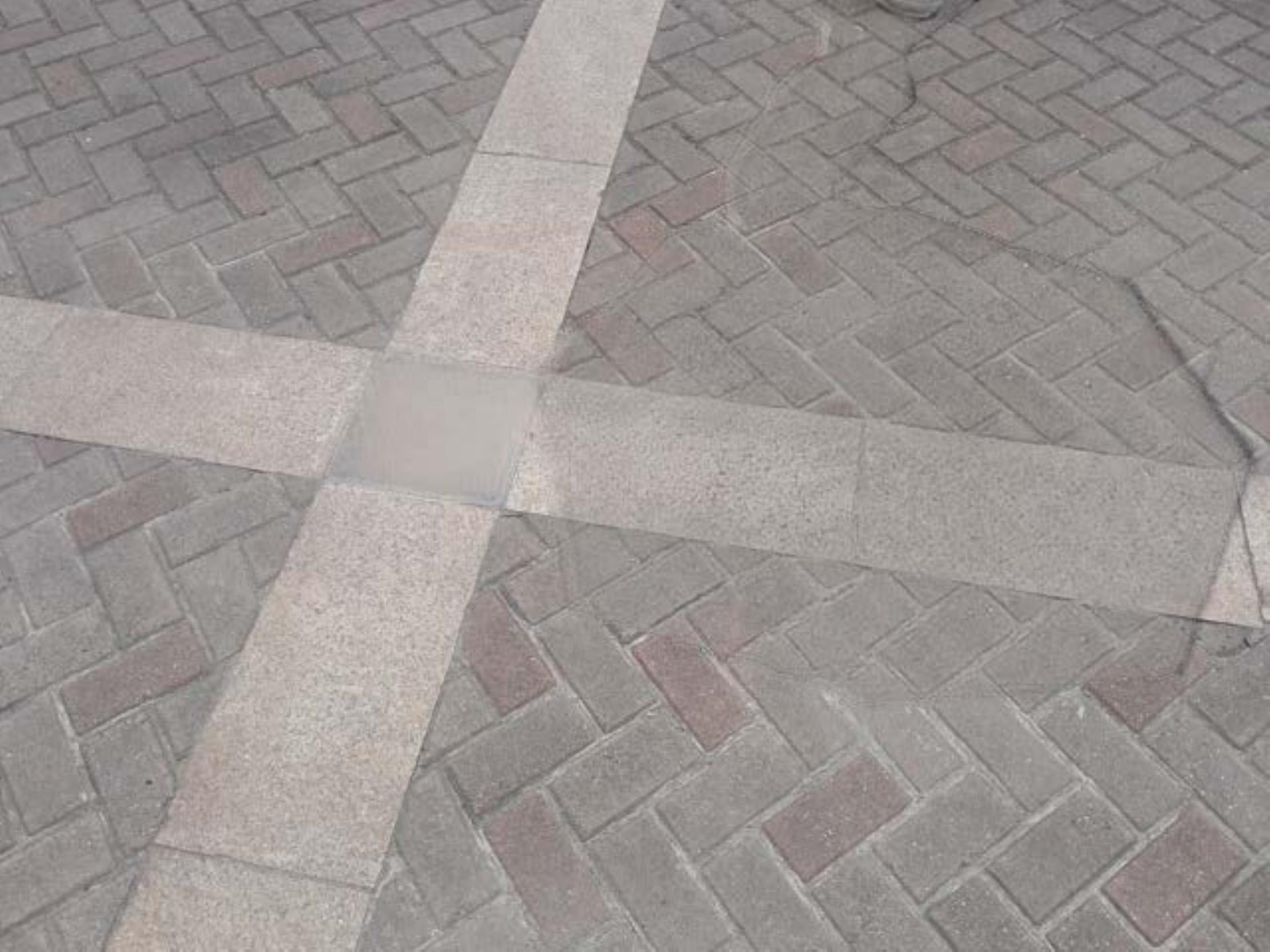}
\caption{G2R}
\end{subfigure}
\begin{subfigure}[t]{0.09\textwidth}
\includegraphics[width=\linewidth, height=\linewidth]{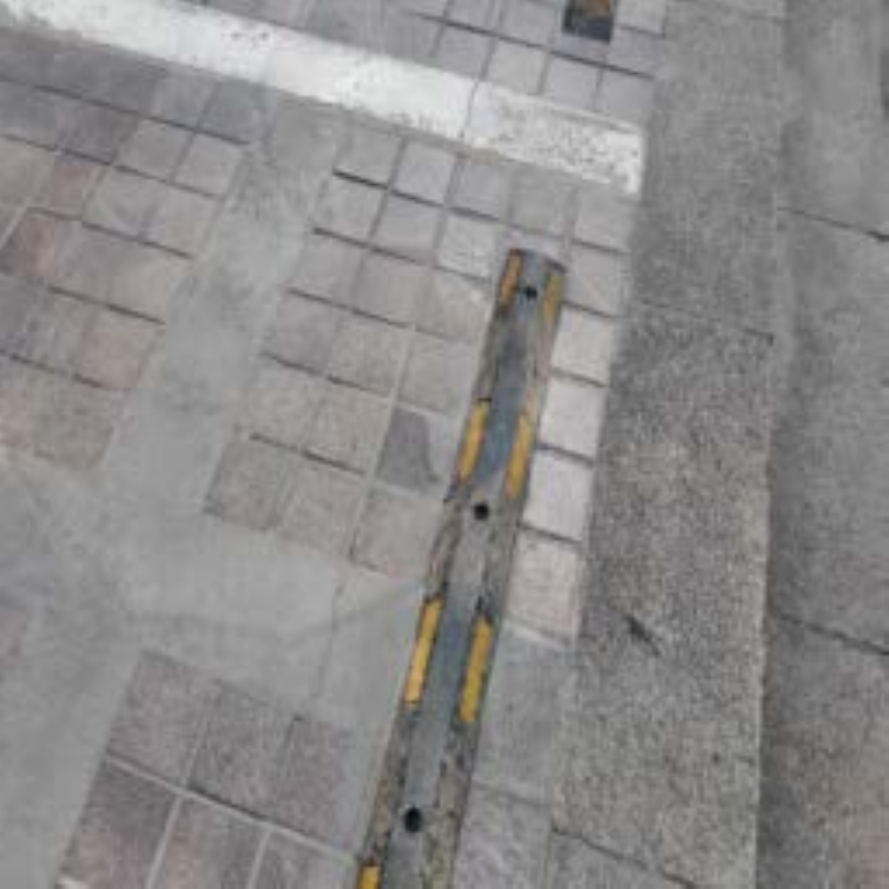} \\
\includegraphics[width=\linewidth, height=\linewidth]{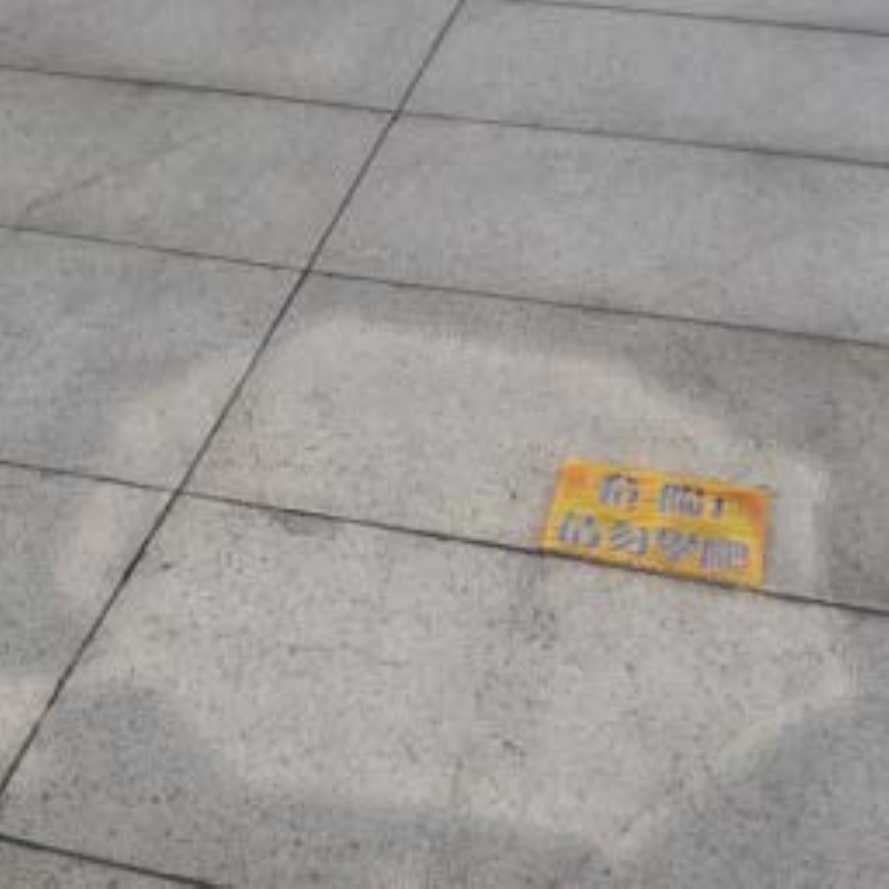} \\
\includegraphics[width=\linewidth, height=\linewidth]{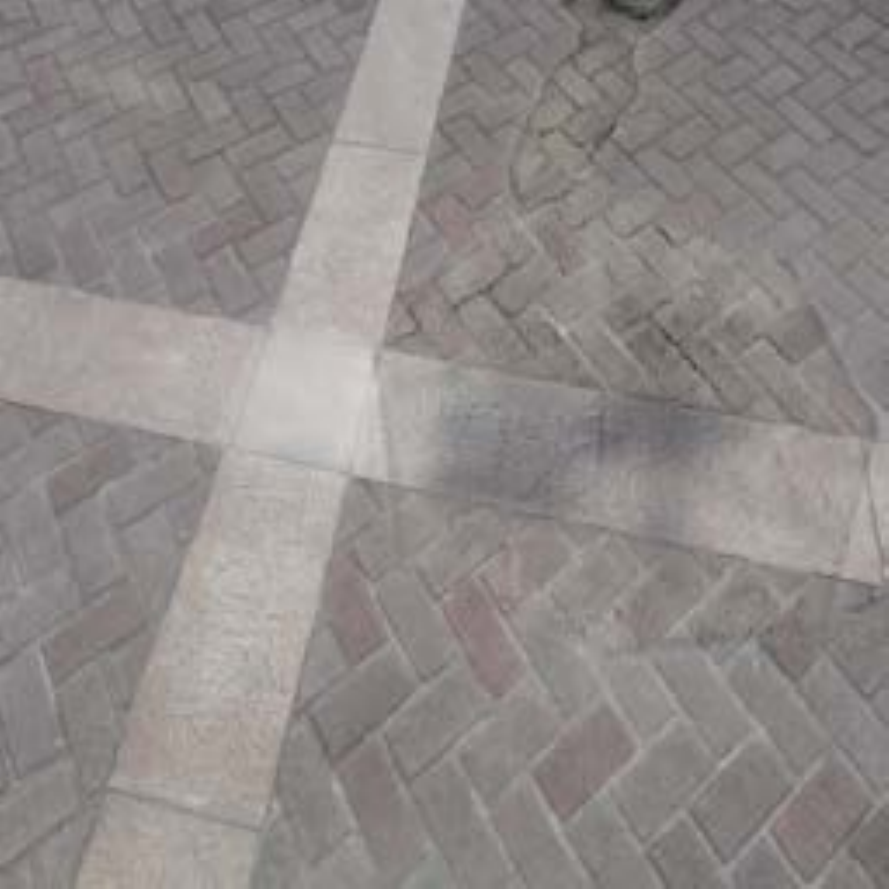}
\caption{DC-SNet}
\end{subfigure}
\begin{subfigure}[t]{0.09\textwidth}
\includegraphics[width=\linewidth, height=\linewidth]{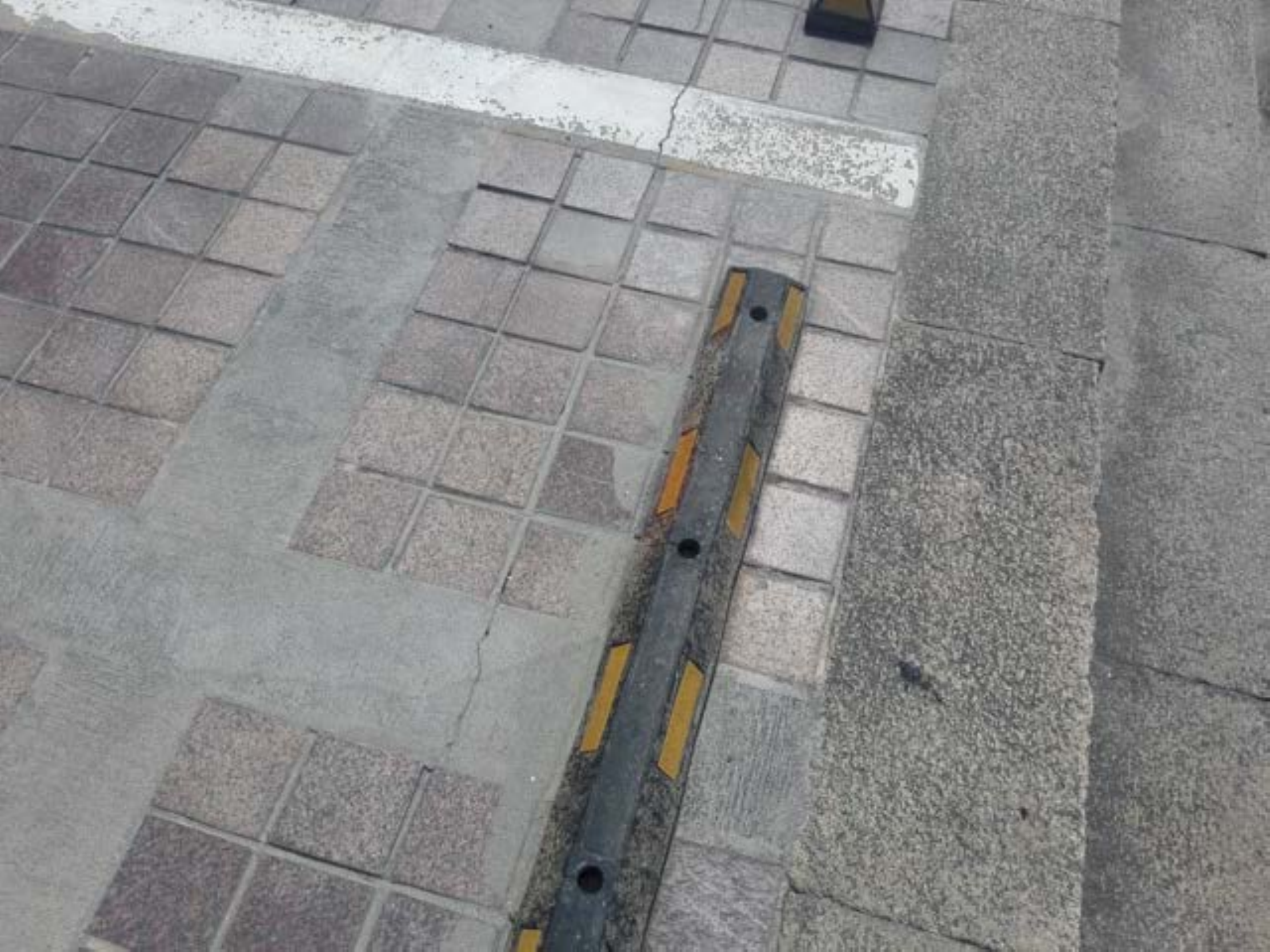} \\
\includegraphics[width=\linewidth, height=\linewidth]{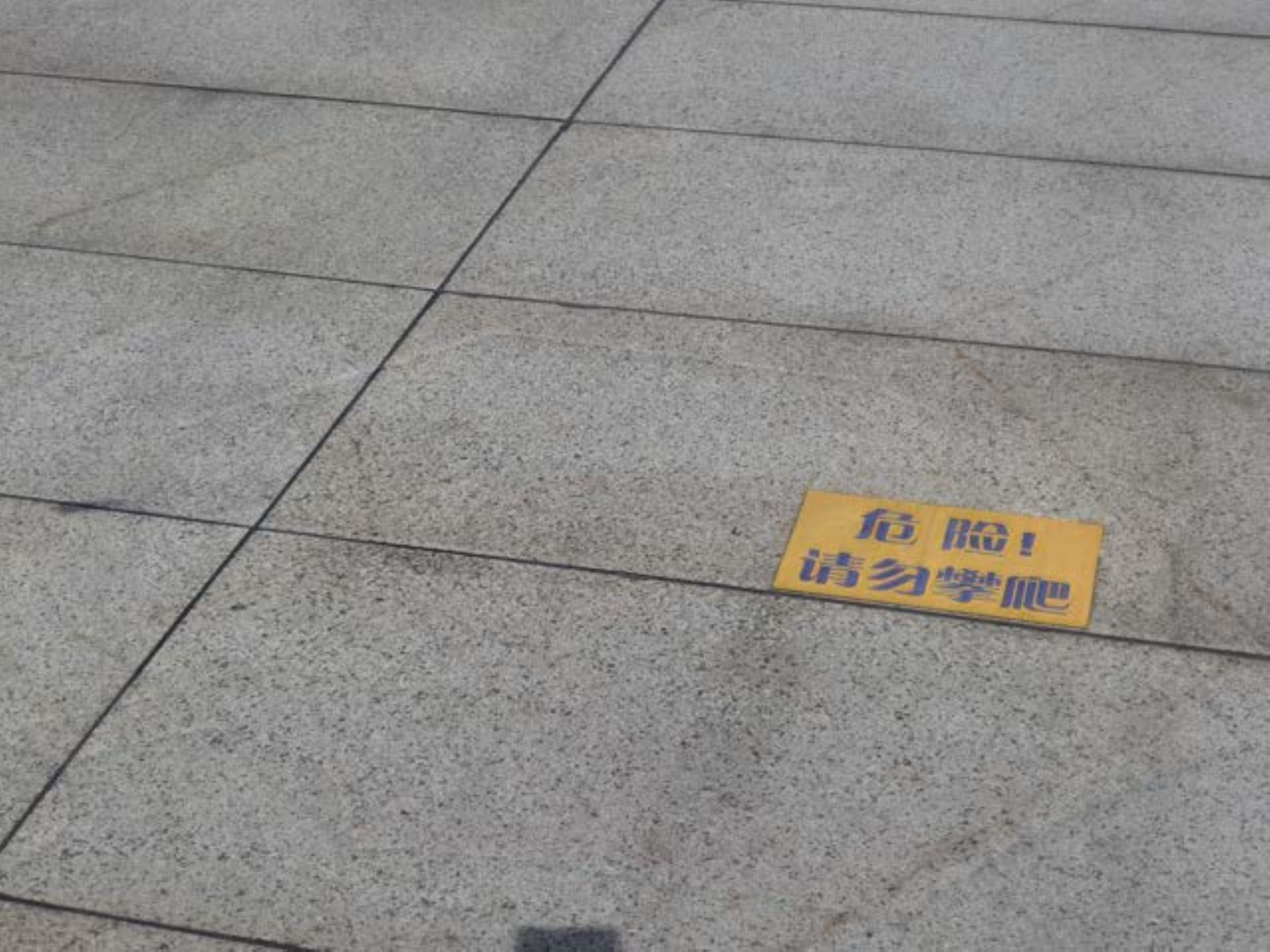} \\
\includegraphics[width=\linewidth, height=\linewidth]{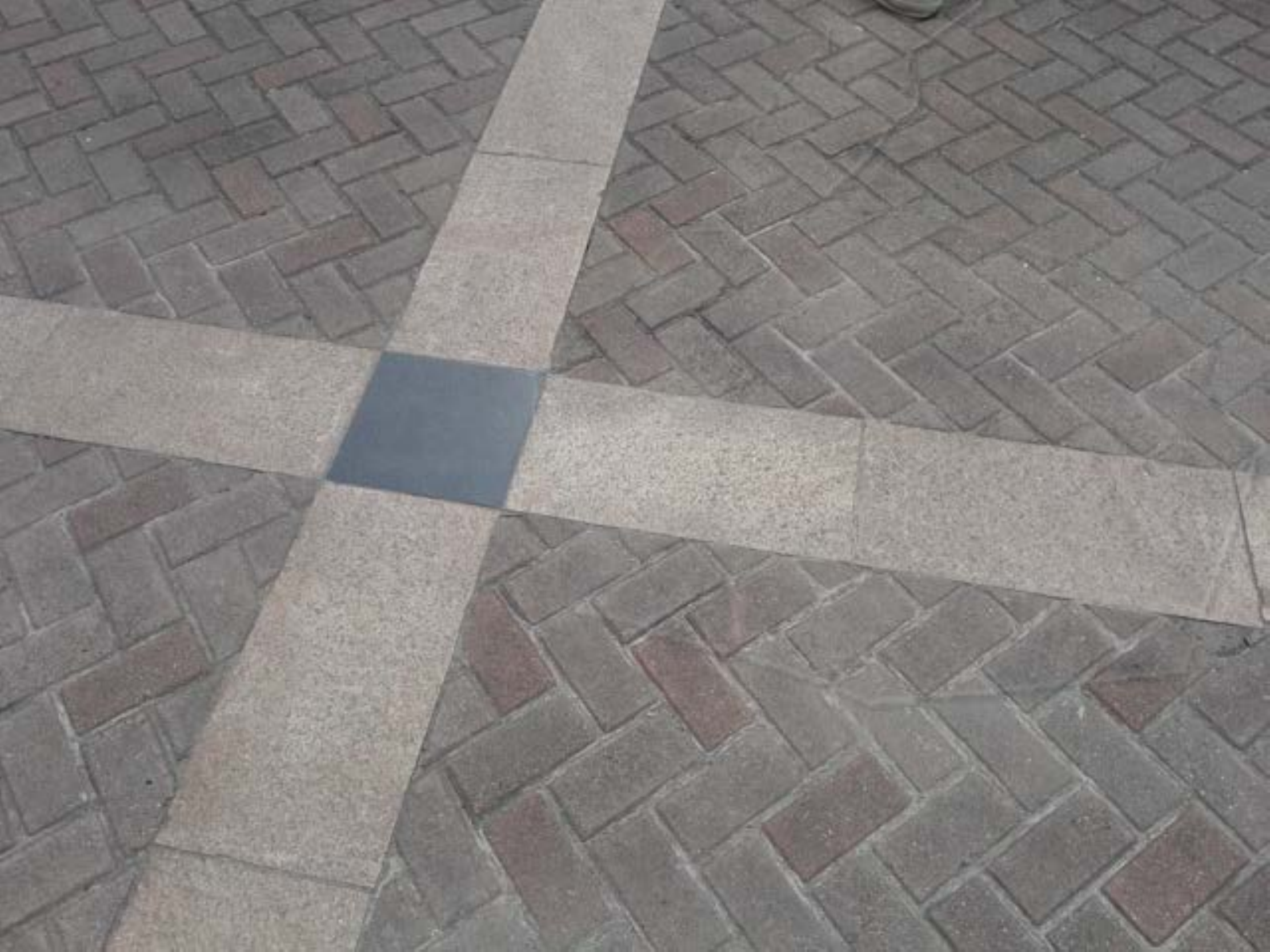}
\caption{Ours}
\end{subfigure}
\begin{subfigure}[t]{0.09\textwidth}
\includegraphics[width=\linewidth, height=\linewidth]{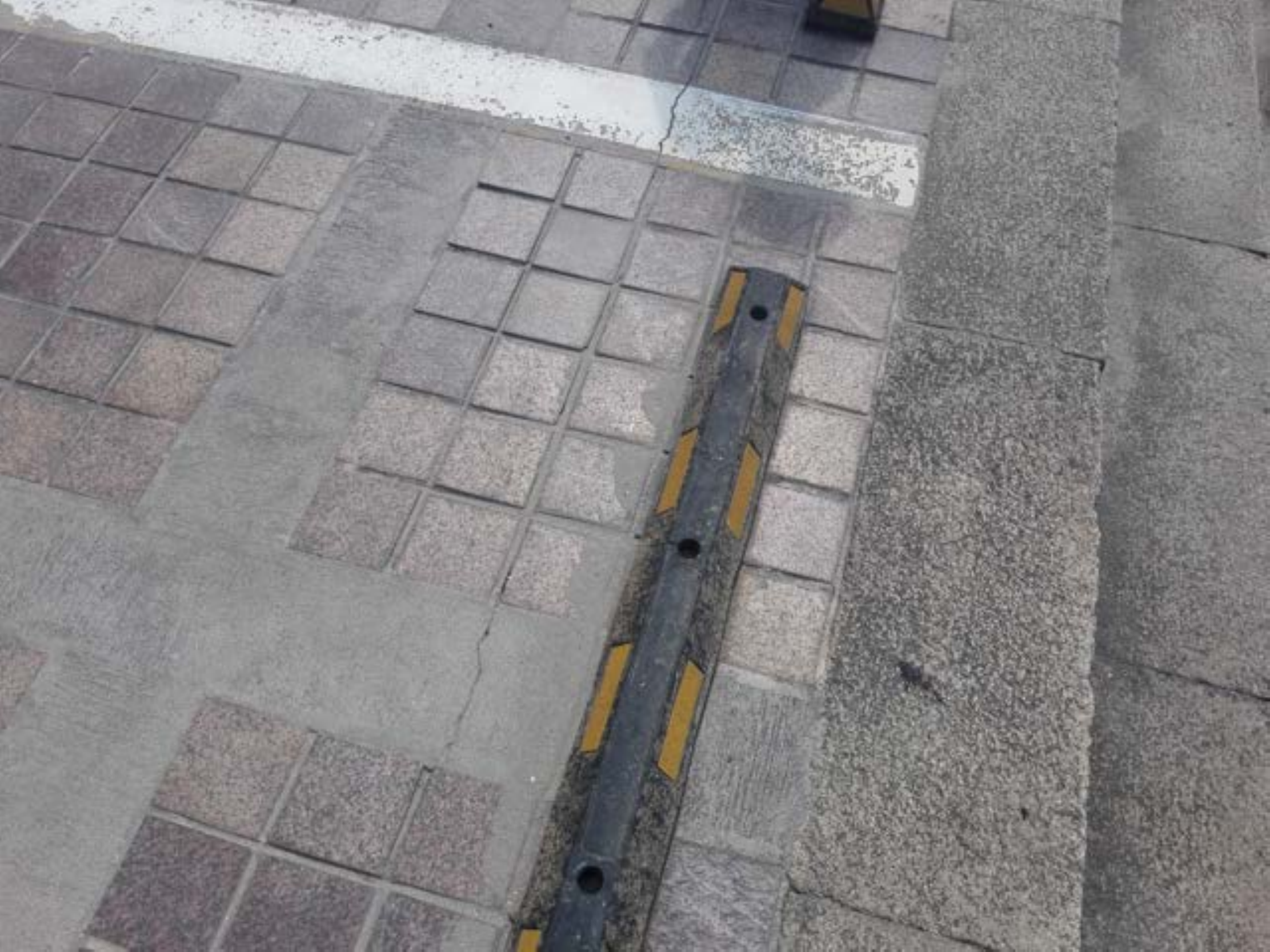} \\
\includegraphics[width=\linewidth, height=\linewidth]{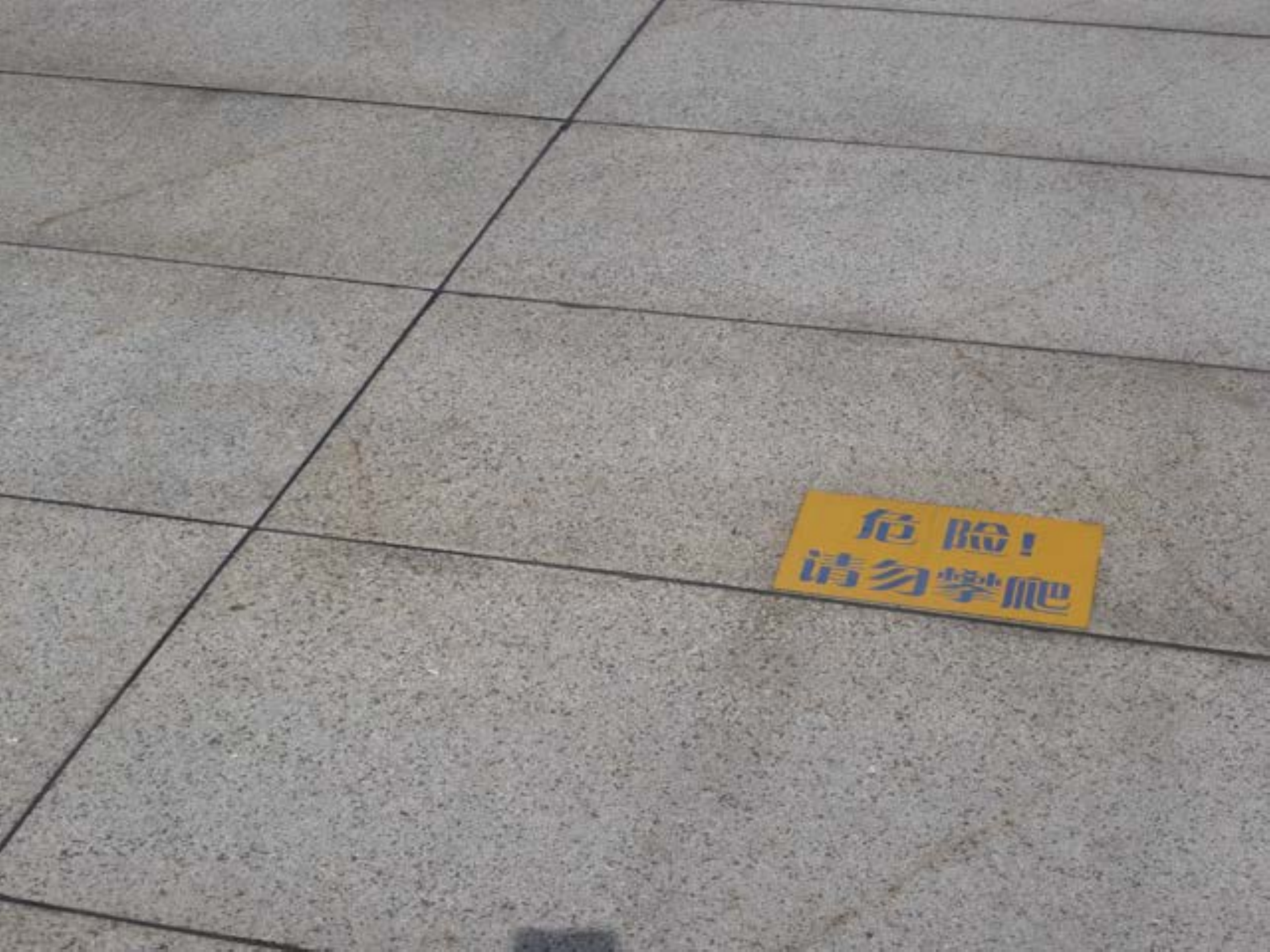} \\
\includegraphics[width=\linewidth, height=\linewidth]{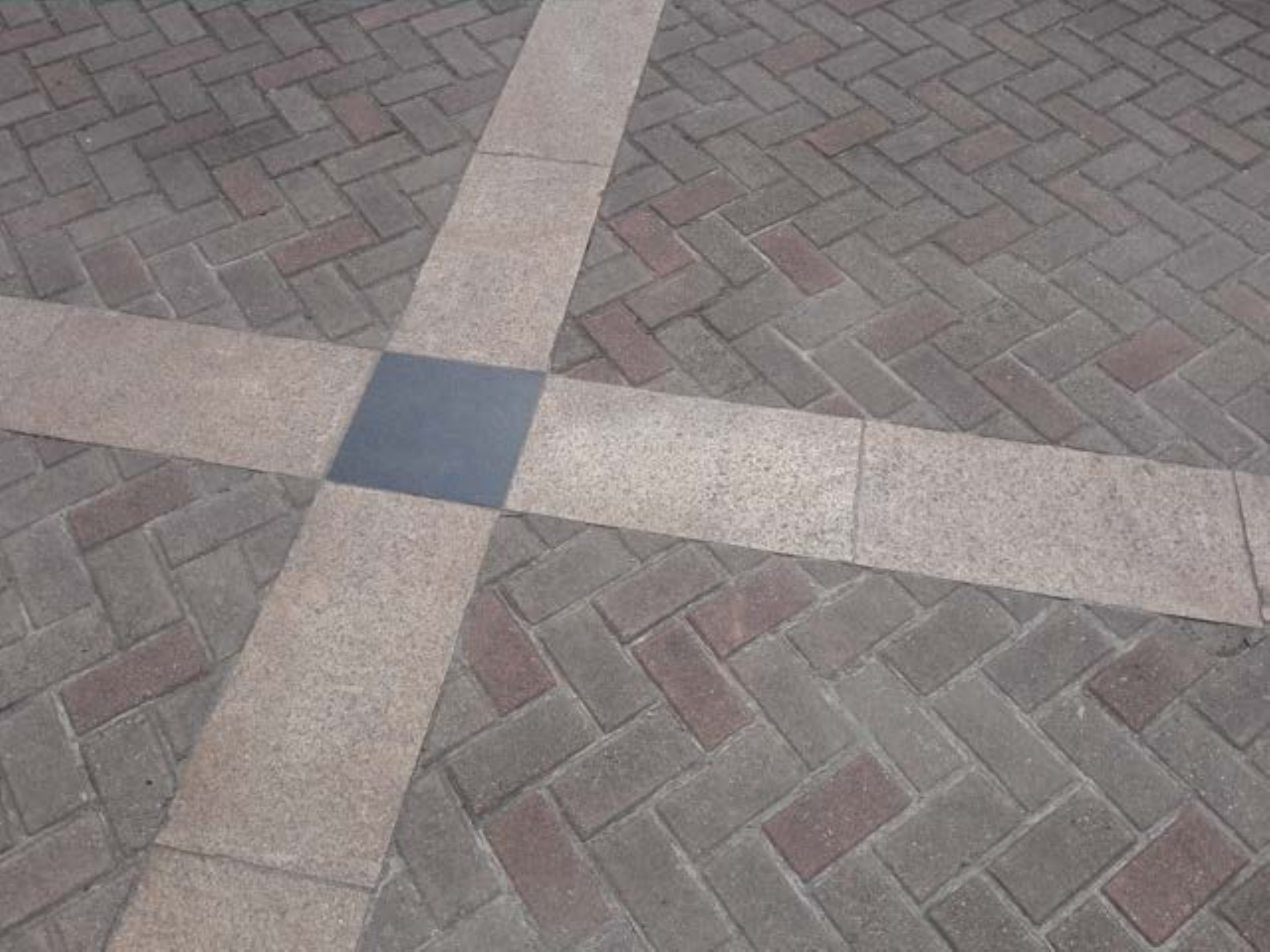}
\caption{GT}
\end{subfigure}
\caption{Visualization of shadow removal results of LAB-Net and current methods including Guo \emph{et al.}~\cite{guo2012paired}, LG-SNet~\cite{liu2021shadow_b},  DHAN~\cite{cun2020towards},
A-E~\cite{fu2021auto}, G2R~\cite{liu2021shadow}and DC-SNet~\cite{jin2021dc}, on ISTD dataset.}
\label{fig:istd_vis}
\end{figure*}

\subsection{Training Objective}
The overall training objective of our LAB-Net consists of three main components including a mean square loss $\mathcal{L}_{\text{mse}}$, a perceptual loss $\mathcal{L}_{\text{percep}}$ and a gradient loss $\mathcal{L}_{\text{grad}}$:

\vspace{-1em}
\begin{align}
    \mathcal{L}_{\text{LAB-Net}} = \mathcal{L}_{\text{mse}} + {\lambda}_1 \mathcal{L}_{\text{percep}} + {\lambda}_2 \mathcal{L}_{\text{grad}}.   
\end{align}
where $\lambda_1$ and $\lambda_2$ are two hyper-parameters, which are respectively set to 10 and 100.

The $\mathcal{L}_{\text{mse}}$ preserves the visual consistency between the generated shadow-free image and its ground truth:

\vspace{-1em}
\begin{align}
    \mathcal{L}_{\text{mse}} = ||I_{\text{free}}, I_{\text{gt}}||_2.
\end{align}

The $\mathcal{L}_{\text{percep}}$ retains the image structure with the aid of a pre-trained VGG feature extractor VGG-16~\cite{simonyan2014very}:

\vspace{-1em}
\begin{align}
    \mathcal{L}_{\text{percep}} = ||VGG(I_{\text{free}}), VGG(I_{\text{gt}})||_1.
\end{align}

The $\mathcal{L}_{\text{grad}}$ measures the gradient differences between the output shadow-free image and its ground-truth:

\vspace{-1em}
\begin{align}
    \mathcal{L}_{\text{grad}} = ||\nabla I_{\text{free}}, \nabla I_{\text{gt}}||_1.
    \label{equation:grad_loss}
\end{align}
where $\nabla$ returns the image gradient.

\section{Experiments}
\subsection{Datasets and Evaluation Metrics}
\textbf{Datasets}. We train and evaluate our LAB-Net on two shadow removal datasets. One is the ISTD dataset ~\cite{wang2018stacked}, which includes 1,330 training triples (shadow image, shadow mask, shadow-free image) and 540 testing triples. The other is the SRD dataset ~\cite{qu2017deshadownet}, which includes 2,680 training pairs (shadow image, shadow-free image) and 408 testing pairs. Since the SRD does not provide corresponding shadow masks, we use public SRD shadow masks from ~\cite{cun2020towards} for training and evaluations.

\textbf{Evaluation Metrics}. 
We evaluate our model performance by computing the root mean square error (RMSE) in the LAB color space between the generated results and ground truth. In addition, we also evaluate the shadow removal effect of various existing methods by calculating the peak signal-to-noise ratio (PSNR) and structural similarity (SSIM) in the RGB color space. For fair comparisons, we directly adopt the MATLAB evaluation codes of ~\cite{fu2021auto}.

\textbf{Implementation Details}. 
We implement our LAB-Net by using PyTorch. Models are trained and evaluated on an ``A100 GPU''. During training, for the ISTD dataset, the input size of images is set to 256$ \times $256, the batch size is set to 2, the training epoch is set to 300, and the downsample size $M$ in the LSA module is set to 256. For the SRD dataset, which contains larger images and is more difficult to converge, the input size of the images is set to 400$ \times $400, the batch size is set to 1, the training epoch is set to 500, and the downsample size $M$ is set to 128. For both datasets, the Adam optimizer with the learning rate of 0.0002 is adopted.

\subsection{Results on ISTD Dataset}
We first compare our results with other state-of-the-art shadow removal methods on the ISTD dataset, including physics-based method~\cite{guo2012paired} and DNNs-based methods~\cite{wang2018stacked,hu2019direction,le2020shadow,cun2020towards,liu2021shadow_b,fu2021auto,liu2021shadow,jin2021dc,zhu2022efficient}. For fair comparisons, results of these methods are provided by the authors or cited from the original paper. Quantitative comparisons are shown in Table.\,\ref{Table:Quant_res_istd}. Our method achieves the best shadow removal performance in shadow regions (S), non-shadow regions (NS), and all the images (ALL). In addition, our method has the smallest amount of network parameters compared to the other methods while maintaining a small computational cost. Specifically, compared to~\cite{zhu2022efficient}, it has the same amount of computational costs as our method, but our method reduces the RMSE value from 8.29 to 6.65 and improves the PSNR value from 36.95 to 37.17, using only 9.3\% of Zhu et al.'s network parameters. We also present the visual comparison results in Fig.\,\ref{fig:istd_vis}. The figure shows that the method based on the physical model cannot effectively remove shadows. While the other DNNs-based methods have color differences between shadow and non-shadow regions. Compared with the previous methods, our method can effectively remove shadows and ensure the color and content consistency of shadow and non-shadow regions.

\begin{figure}[!t]
    \centering
    \begin{subfigure}[ht]{0.17\linewidth}
        \includegraphics[width=\linewidth, height=\linewidth]{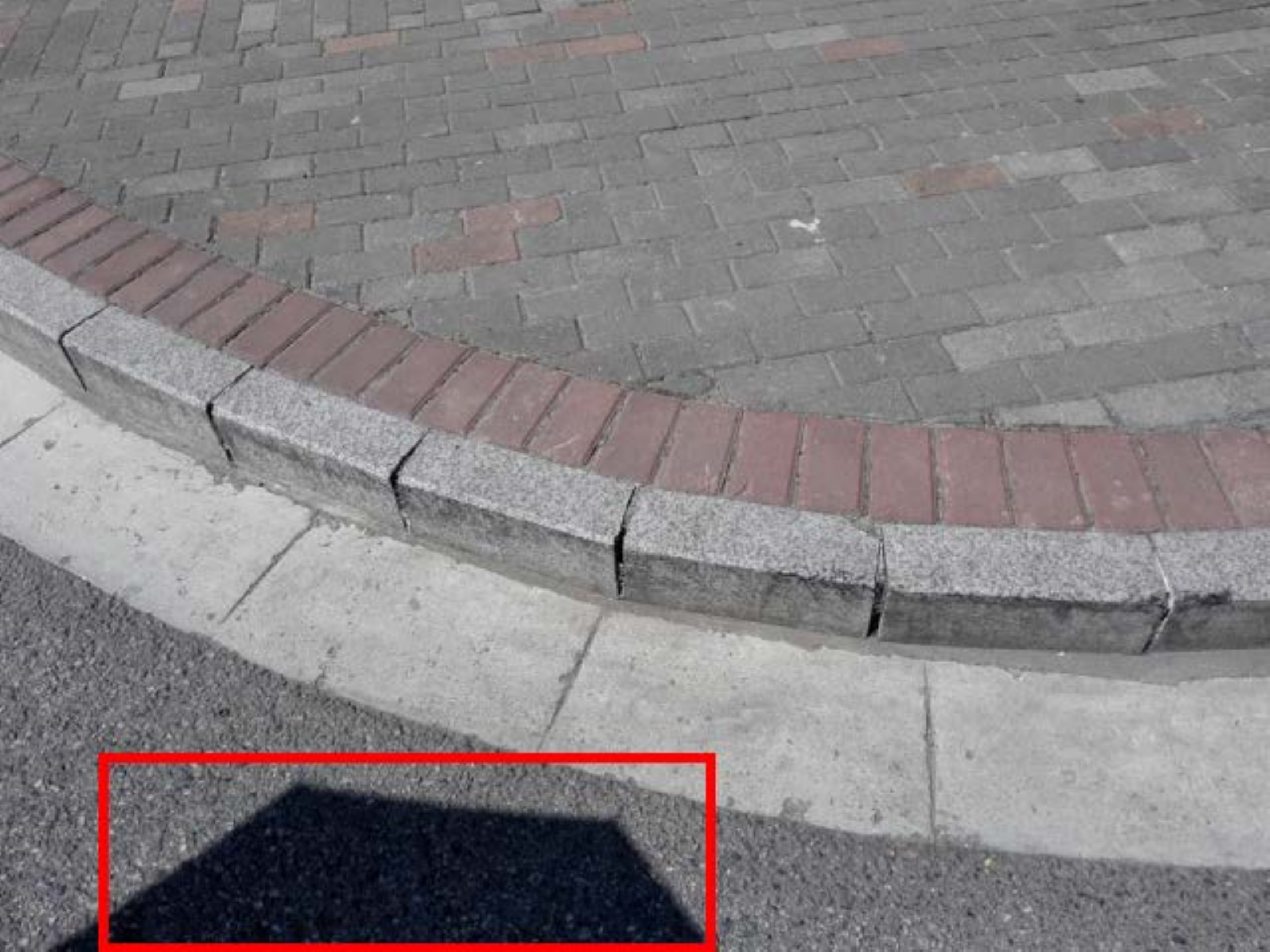}\\
        \includegraphics[width=\linewidth, height=0.4\linewidth]{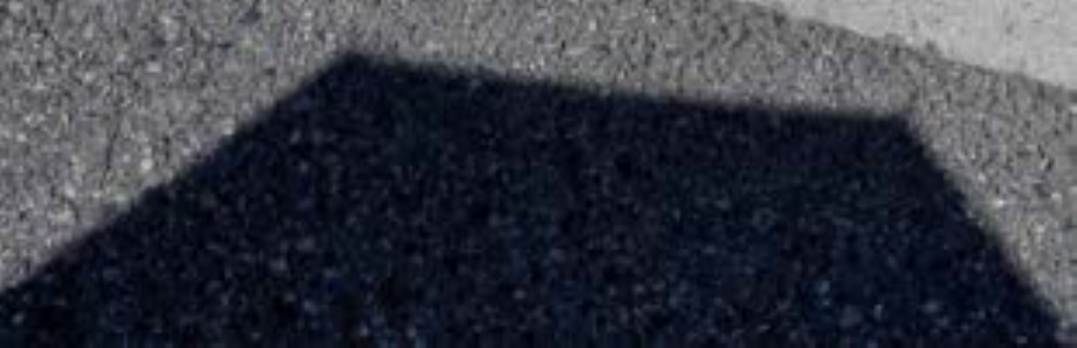}
    \caption{\scriptsize{Input}}
    \end{subfigure}
    \begin{subfigure}[ht]{0.17\linewidth}
        \includegraphics[width=\linewidth, height=\linewidth]{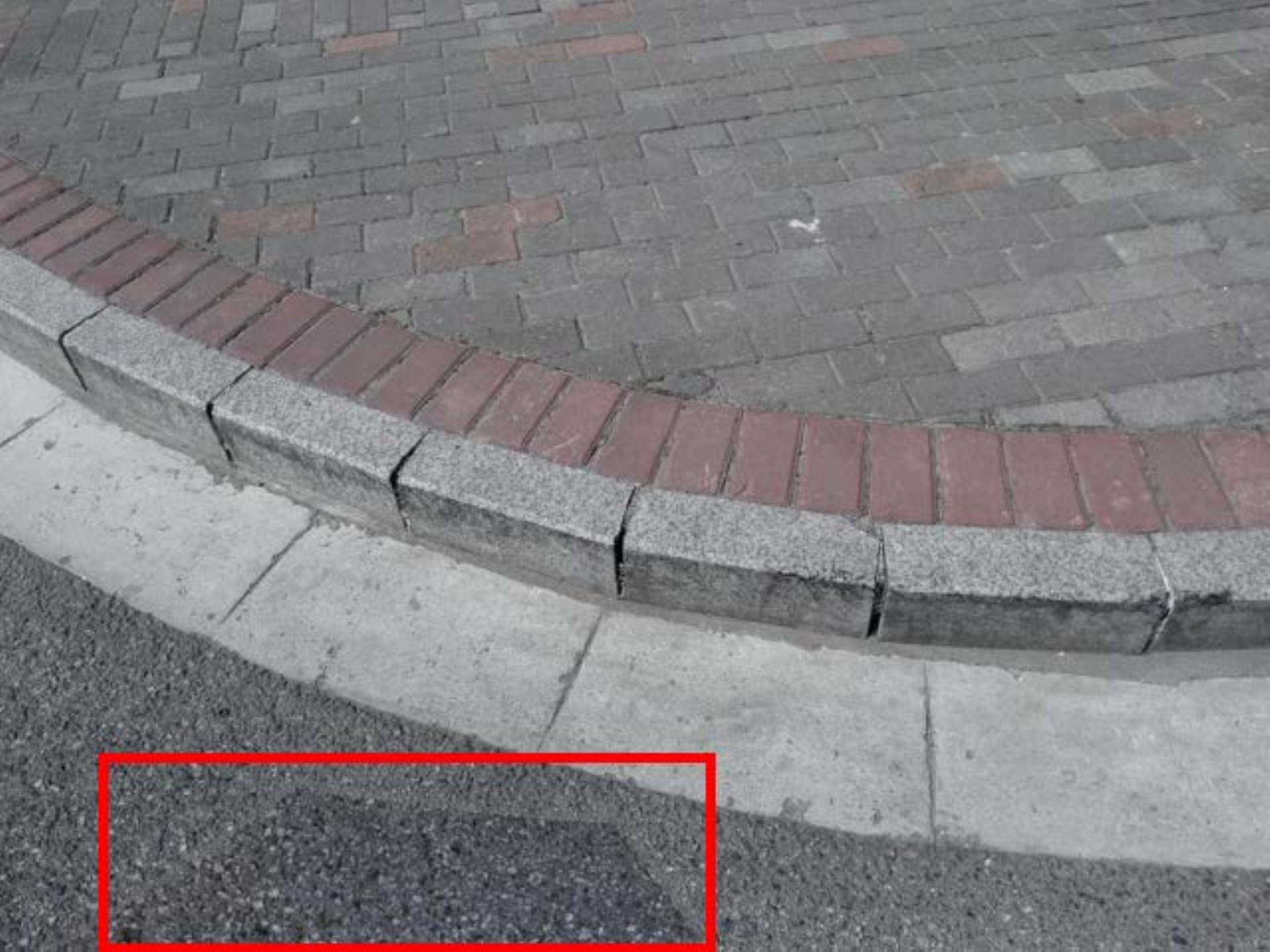} \\
        \includegraphics[width=\linewidth, height=0.4\linewidth]{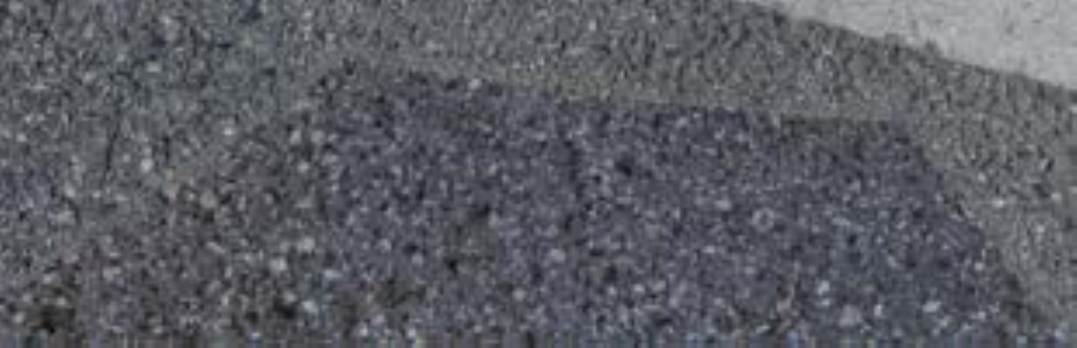}
    \caption{\scriptsize{Base}}
    \end{subfigure}
    \begin{subfigure}[ht]{0.17\linewidth}
        \includegraphics[width=\linewidth, height=\linewidth]{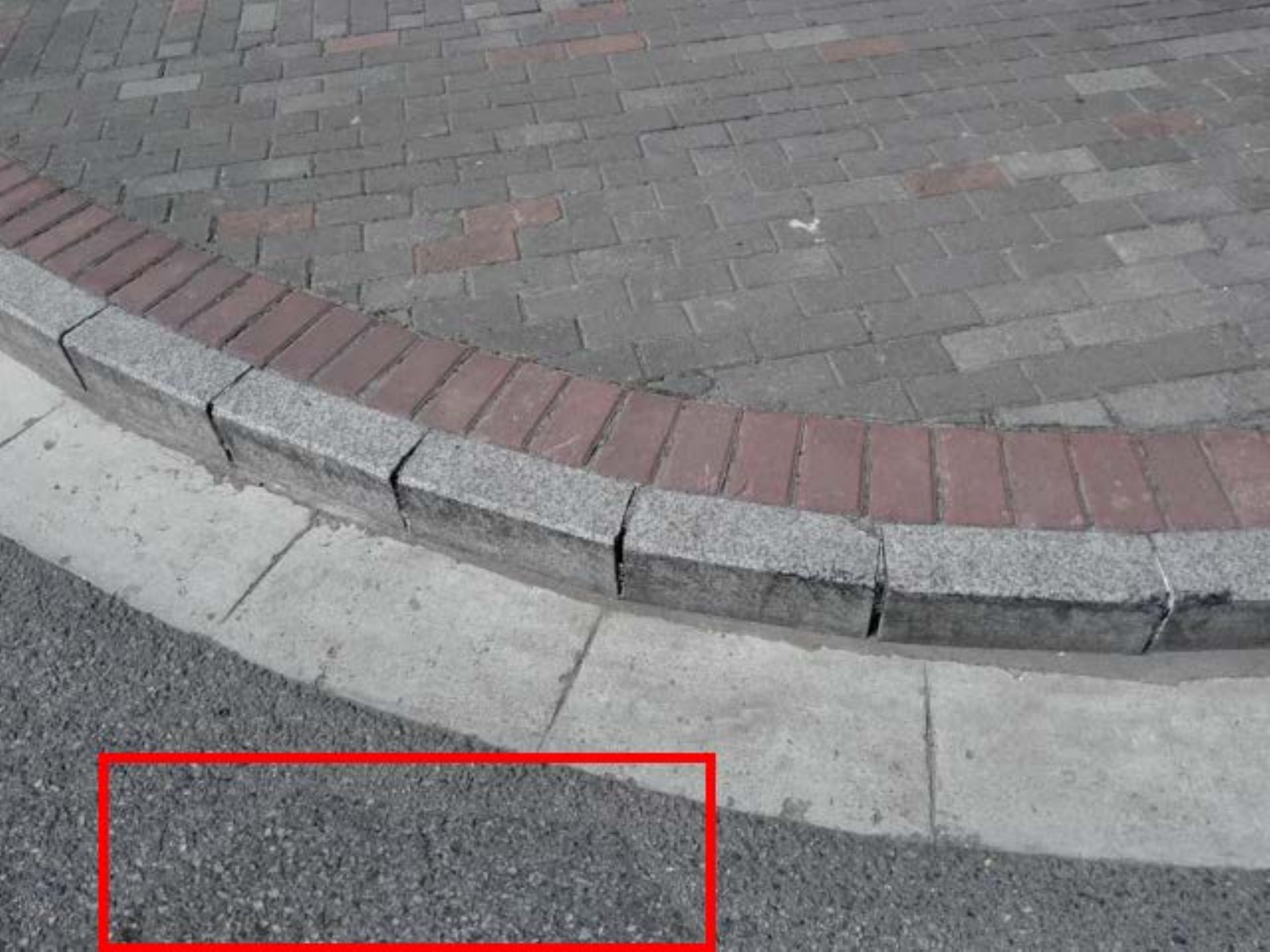} \\
        \includegraphics[width=\linewidth, height=0.4\linewidth]{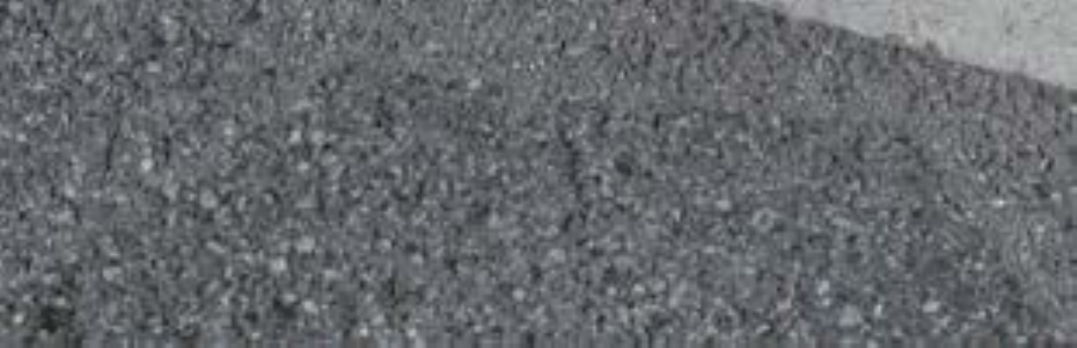}
    \caption{\tiny{Base + ECA}}
    \end{subfigure}
    \begin{subfigure}[ht]{0.17\linewidth}
        \includegraphics[width=\linewidth, height=\linewidth]{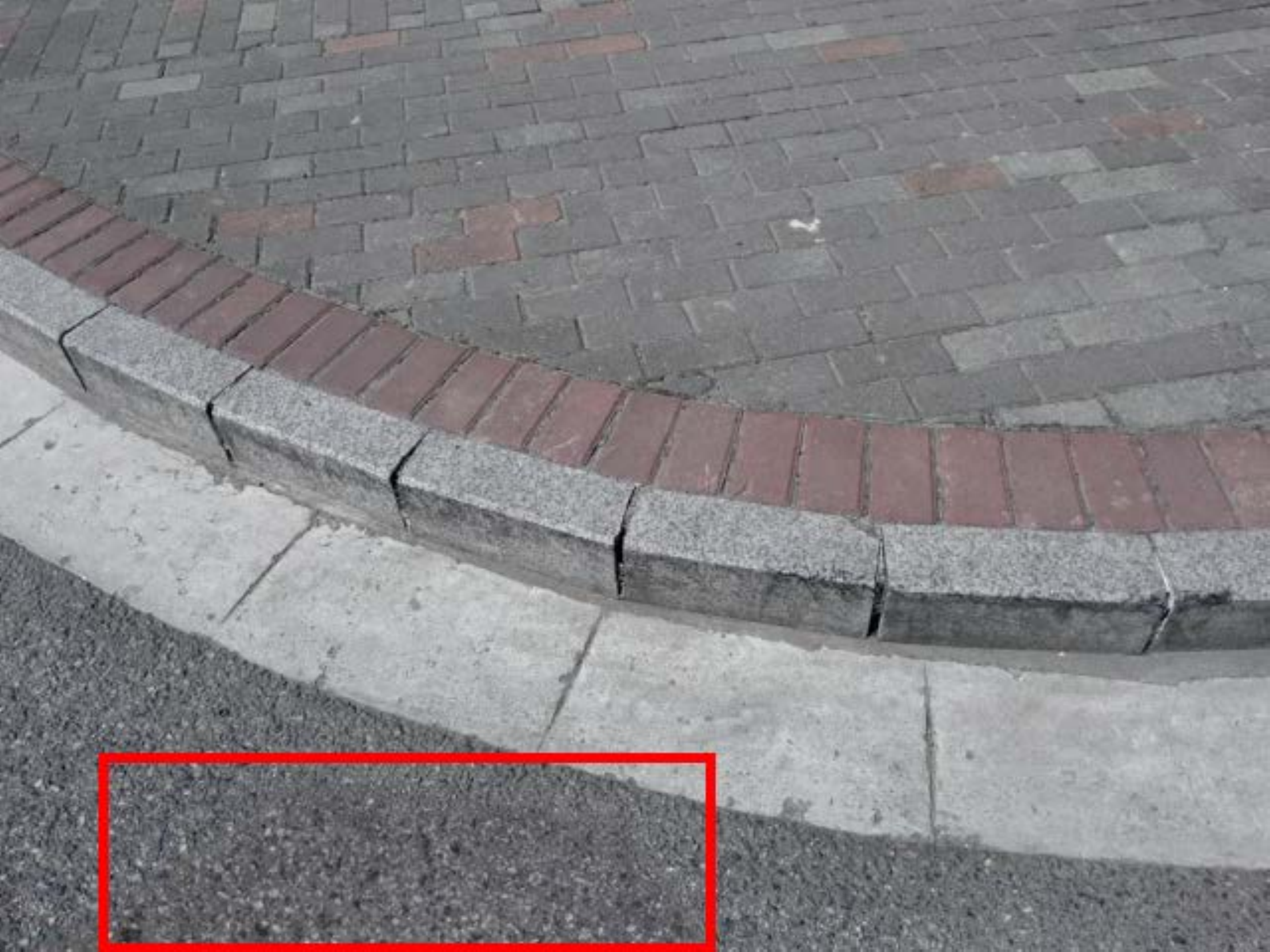} \\
        \includegraphics[width=\linewidth, height=0.4\linewidth]{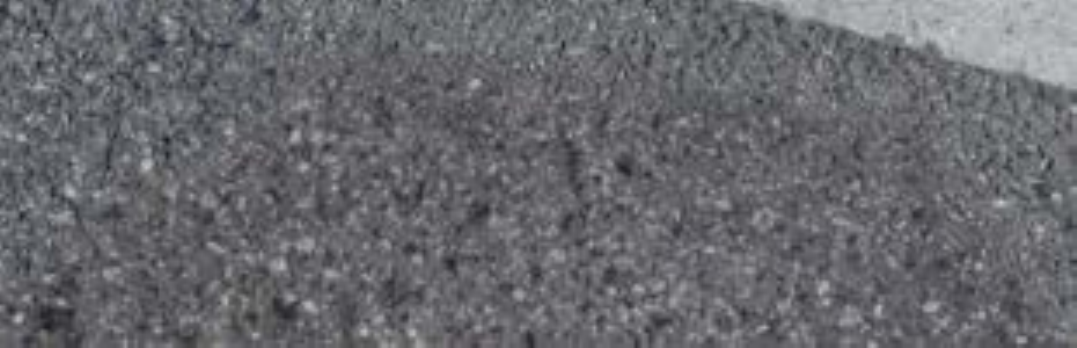}
    \caption{\tiny{Base + LSA}}
    \end{subfigure}
    \begin{subfigure}[ht]{0.17\linewidth}
        \includegraphics[width=\linewidth, height=\linewidth]{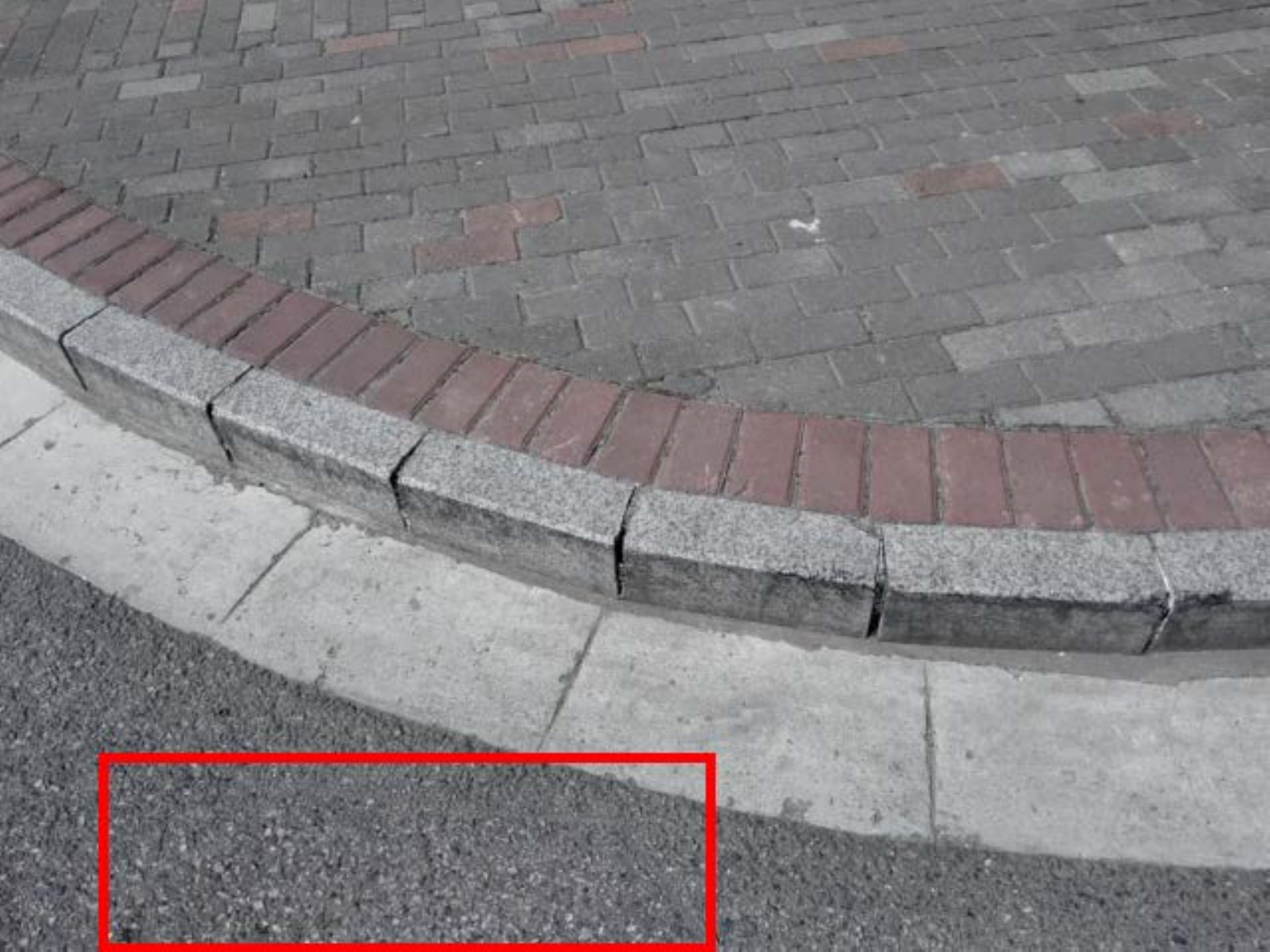} \\
        \includegraphics[width=\linewidth, height=0.4\linewidth]{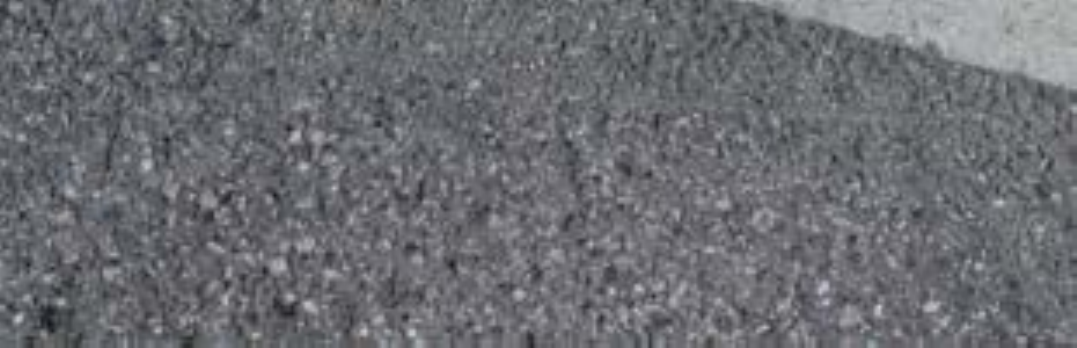}
    \caption{\scriptsize{Ours}}
    \end{subfigure}
    \caption{Qualitative comparison of shadow removal results among different combinations of our methods.}
    \label{fig:ablation modules}
\end{figure}

\subsection{Results on SRD Dataset}
We also compare our results with other state-of-the-art shadow removal methods ~\cite{hu2019direction,cun2020towards,fu2021auto,jin2021dc,zhu2022efficient} on the SRD dataset. The results as shown in Table.\,\ref{Table:Quant_res_srd} indicate that our method also achieves the best performance with the lowest RMSE value and highest PSNR value. Compared with EMDN, the RMSE value is reduced from 7.44 to 6.56, and the PSNR value is improved from 34.94 to 35.71.

\begin{table*}[!t]
\small
\centering
\caption{Performance comparison on SRD dataset.}
\setlength{\tabcolsep}{4pt}
\begin{tabular}{c|ccc|ccc|ccc}
\hline
\multicolumn{1}{c|}{\multirow{2}{*}{Methods}} &
\multicolumn{3}{c|}{Shadow Region (S)} & \multicolumn{3}{c|}{Non-Shadow Region (NS)} & \multicolumn{3}{c}{All image (ALL)} \\ \cline{2-10} 
\multicolumn{1}{c|}{}  &RMSE $\downarrow$ & PSNR $\uparrow$ & SSIM $\uparrow$ & RMSE $\downarrow$ & PSNR $\uparrow$ & SSIM $\uparrow$ & RMSE $\downarrow$ & PSNR $\uparrow$ & SSIM $\uparrow$ \\ \hline
\multicolumn{1}{c|}{Input Image}  &36.62   &18.98   &0.8720   &4.54   &31.68   &0.9815   &13.83   &18.26   &0.8372   \\ \hline
\multicolumn{1}{c|}{DSC} 
&19.35  &26.46  &0.9033  &15.76  &24.80  &0.7740  &16.77  &21.64  &0.6566  \\
\multicolumn{1}{c|}{DHAN} 
&\underline{6.94}  &33.84  &\underline{0.9797}  &\textbf{3.63}  &35.06  &\textbf{0.9850}  &\underline{4.61}  &30.74 &\textbf{0.9577}  \\
\multicolumn{1}{c|}{Auto-Exposure} 
&8.33  &32.44  &0.9684  &5.48  &30.84  &0.9504  &6.25  &27.97  &0.9016  \\
\multicolumn{1}{c|}{DC-Shadownet}  
&7.53  &33.40  &0.9743  &3.91  &34.95  &\underline{0.9831}  &4.94  &30.56  &0.9473  \\
\multicolumn{1}{c|}{EMDN} 
&7.44  &\underline{34.94}  &0.9797  &\underline{3.74}  &\underline{35.85}  &0.9819  &4.79  &\underline{31.72}  &0.9523  \\
\hline
\multicolumn{1}{c|}{Ours} &\textbf{6.56}  &\textbf{35.71}  &\textbf{0.9818}  &3.77  &\textbf{36.50}  &0.9813  &\textbf{4.60}  &\textbf{32.22}  &\underline{0.9554}  \\ \hline
\end{tabular}
\label{Table:Quant_res_srd}
\normalsize
\end{table*}

\begin{table}[!t]
\small
\centering
    \caption{Efficacy of different Modules in our method.}
    \begin{tabular}{ccc|ccc}
    \hline
    \multicolumn{3}{c|}{Method} & \multicolumn{3}{c}{RMSE $\downarrow$} \\ \hline
    \multicolumn{1}{c|}{Baseline} & \multicolumn{1}{c|}{ECA} & \multicolumn{1}{c|}{LSA} & \multicolumn{1}{c|}{S} & \multicolumn{1}{c|}{NS} & \multicolumn{1}{c}{All} \\ \hline
    \checkmark & \multicolumn{1}{c}{} & \multicolumn{1}{c|}{} & 7.39 & \multicolumn{1}{c}{5.23} & \multicolumn{1}{c}{5.56} \\
    \multicolumn{1}{c}{\checkmark} & \checkmark &  & \multicolumn{1}{c}{6.74} & 4.74 & 5.07 \\
    \multicolumn{1}{c}{\checkmark} &  &\checkmark  & \multicolumn{1}{c}{7.21} & 4.95 & 5.32 \\
    \multicolumn{1}{c}{\checkmark} & \checkmark & \checkmark & \multicolumn{1}{c}{6.65} & 4.49 & 4.84 \\ \hline
    \end{tabular}
    \label{Table:effective}
    \normalsize
\end{table}

\subsection{Ablation Study}
We conduct ablation studies on the ISTD dataset as an example to analyze the influences of the various modules proposed in this paper.

\textbf{Component Effect}.
In Table.\,\ref{Table:effective}, the effectiveness of different modules in our method can be verified by using the ECA module and the LSA module individually. Our proposed two-branch network is used as baseline. After adding the ECA module or the LSA module, the shadow removal performance has improved significantly. To demonstrate the effectiveness of our proposed module, the visualization results are shown in Fig.\,\ref{fig:ablation modules}.

\textbf{Network Structure}.
In this work, we discuss that the two-branch network structure based on the LAB color space is more suitable for the shadow removal task. We conduct experiments to prove the validity of the baseline. We merge the two-branch structure in the network into a single branch, and optimize the LAB as a whole, denoting it as LAB$_{\text{together}}$. In addition, we also train the single-branch network in the RGB color space, denoting it as RGB$_{\text{together}}$. Table.\,\ref{Table:Ablation} clearly shows that the LAB color space is better than the RGB color space for the shadow removal task. The independent optimization of L, A \& B is also better than the joint optimization.

\textbf{Network Width \emph{w.r.t.} Dilation Rate}.
In order to reasonably reduce the parameters of the network, in our Basic Block, we reserve more channels for the dilated convolution with a large dilation rate. For the dilated convolution under the same level feature, the number of channels is reserved as 16-32-48 according to the size of the dilation rate. To demonstrate the effectiveness of this reservation strategy, We also try other channel reservation strategies, such as 48-32-16 and 32-32-32. The results shown in Table.\,\ref{Table:Ablation} indicate that our channel reservation strategy achieves the best performance after reducing parameters.

\textbf{ECA module}.
In the ECA module, we use the Laplacian filter that can preserve more spatial details of the feature. We also conduct experiments using the Sobel filter and Global Average Pooling (GAP) method. The results are shown in Table.\,\ref{Table:Ablation}, which illustrates that the Laplacian filter is more suitable for the shadow removal task.

\begin{table}[!t]
\small
\centering
\caption{Ablation studies of different variants of our method.}
\begin{tabular}{c|ccc}
\hline
\multicolumn{1}{c|}{\multirow{2}{*}{Methods}} & \multicolumn{3}{c}{RMSE $\downarrow$}\\ \cline{2-4}
\multicolumn{1}{c}{} &
\multicolumn{1}{|c|}{S} & \multicolumn{1}{c|}{NS} &  \multicolumn{1}{c}{ALL} \\ \hline
\multicolumn{4}{c}{\textbf{Network Structure}}\\ \hline
RGB$_{\text{together}}$ & 8.34 & 4.88 & 5.47 \\
LAB$_{\text{together}}$ & 8.11 & 5.06 & 5.57 \\
LAB-Net(Ours) & 7.39 & 5.23 & 5.56 \\ \hline
\multicolumn{4}{c}{\textbf{Network Width \emph{w.r.t.} Dilation Rate}}\\ \hline
64-64-64 &6.38 &4.59 &4.86 \\ 
\hline
48-32-16 & 7.15& 4.66 & 5.06 \\
32-32-32 & 6.96& 4.55 & 4.93 \\
16-32-48(Ours) & 6.65 & 4.49 & 4.84 \\\hline
\multicolumn{4}{c}{\textbf{ECA module}}\\ \hline
GAP & 7.19& 5.13& 5.43 \\
Std[Sobel] & 7.20& 4.94 & 5.28 \\
Std[Laplacian] (Ours) & 6.74 & 4.74 & 5.07 \\ \hline
\multicolumn{4}{c}{\textbf{LSA module}}\\ \hline
Whole & 7.75& 4.72 & 5.24 \\
Local(Ours) & 6.65 & 4.49 & 4.84 \\\hline
no downsample & 6.22& 4.25 & 4.55 \\
downsample$_{256 \times 256}$(Ours) & 6.65 & 4.49 & 4.84 \\ \hline
\end{tabular}
\label{Table:Ablation}
\normalsize
\end{table}

\textbf{LSA module}.
1) Non-shadow regions.
In the LSA module, for the selection of the non-shadow regions, we use the local non-shadow regions around the boundary of shadows. We also compare the results with whole non-shadow regions.  Table.\,\ref{Table:Ablation} shows that the effect of using local non-shadow regions is better since locality can avoid the influence of irrelevant information. Simultaneously, the local non-shadow regions also use less computational costs.

2) Downsampling size.
In the LSA module, we reduce the computational cost of the LSA module by using a downsampling operation on the input feature to change its size to 256$\times$256. We also conduct experiments without downsampling operations. The results shown in Table.\,\ref{Table:Ablation} indicate that no downsample operation can greatly improve the performance of shadow removal. Nevertheless, we still adopt the downsampling strategy to pursue a better computation-performance trade-off in this paper. Consequently, with the great reduction in computational costs, our LAB-Net also leads to better performance as discussed earlier. Users can flexibly adjust the downsampling size according to the available resources of their computational platforms.

\section{Conclusion}
We presented a novel lightweight network that processes shadow images on the LAB color space in this paper. The network was organized as a two-branch structure to process luminance channel L and color channels A \& B in parallel. Then, we constructed the Basic Block of the proposed network through a parallel structure. Moreover, by observing shadow images, we found that the local non-shadow regions around the boundary of the shadow regions were similar to the shadow regions. Thus, we designed a local spatial attention module that uses the information of local non-shadow regions to guide the network to better remove shadows. Finally, our experiments demonstrated the effectiveness of our method, and our method achieved the best performance with less complexity. Given these merits, our proposed network can be used to handle other artificial intelligence tasks related to image luminance, such as highlight removal \cite{fu2021multi}, which will be considered as our future work.


{\small
\bibliographystyle{ieee_fullname}
\bibliography{egbib}
}

\end{document}